\documentclass[11pt,letterpaper]{article}
\usepackage[T1]{fontenc}
\usepackage[utf8]{inputenc}
\usepackage{arxiv}
\usepackage{newtxtext}
\usepackage{graphicx,amsmath,amssymb,booktabs,longtable,array,calc,xurl,needspace,placeins,textcomp}
\newcommand{\ul}[1]{\underline{#1}}

\usepackage{newtxmath}
\usepackage{microtype}
\usepackage{hyperref}
\hypersetup{hidelinks,pdftitle={Clinician use of language models diverges from how the models are evaluated}}
\makeatletter
\renewcommand{\normalsize}{\@setfontsize\normalsize{11}{13.5}}
\renewcommand{\small}{\@setfontsize\small{9.5}{11.5}}
\normalsize
\makeatother
\begin{document}

\begingroup\centering\setlength{\parskip}{0pt}
{\fontsize{18}{21}\selectfont\bfseries Clinician use of language models diverges from how the models are evaluated\par}
\vspace{12pt}{\fontsize{11}{14}\selectfont \mbox{Krithik Vishwanath BS\textsuperscript{1}}, \mbox{Haitong Lin MS\textsuperscript{2}}, \mbox{Anton Alyakin MSE\textsuperscript{1,3,4}}, \mbox{Jin Vivian Lee MD MS\textsuperscript{1,4}}, \mbox{D. Brock Hewitt MD MPH\textsuperscript{5}}, \mbox{Jie J. Yao MD\textsuperscript{6}}, \mbox{William Robert Small MD MBA\textsuperscript{7,8}}, \mbox{Hammad A. Khan MD\textsuperscript{1}}, \mbox{Cordelia Orillac MD\textsuperscript{1}}, \mbox{Aakaash Varma MD\textsuperscript{9}}, \mbox{Brandon Ye MS\textsuperscript{1,10}}, \mbox{Daniel Alexander Alber MD\textsuperscript{11}}, \mbox{Gustavo Stolovitzky PhD\textsuperscript{12,13}}, \mbox{Batia Wiesenfeld PhD\textsuperscript{14}}, \mbox{Oded Nov PhD\textsuperscript{2}}, \mbox{Wei Wu BS\textsuperscript{15,16}}, \mbox{Kang Zhang MD PhD\textsuperscript{15}}, \mbox{Yindalon Aphinyanaphongs MD PhD\textsuperscript{7,8,17}}, \mbox{Tim Requarth PhD\textsuperscript{18\ensuremath{\ddagger}}}, \mbox{Eric Karl Oermann MD\textsuperscript{1,4,19--21\ensuremath{\ddagger}} \& The International Digital Twin Consortium in Healthcare and Medicine}\par}
\vspace{4pt}{\fontsize{9}{11}\selectfont \ensuremath{\ddagger} - equal co-senior authorship\par}
\vspace{9pt}\fontsize{8.5}{10.5}\selectfont
\textsuperscript{1}Department of Neurosurgery, NYU Langone Health, New York, NY, USA\par
\textsuperscript{2}Department of Technology Management, NYU Tandon School of Engineering, New York University, New York, NY, USA\par
\textsuperscript{3}Washington University School of Medicine, St. Louis, MO, USA\par
\textsuperscript{4}Global AI Frontier Lab, New York University, New York, NY, USA\par
\textsuperscript{5}Department of Surgery, NYU Langone Health, New York, NY, USA\par
\textsuperscript{6}Department of Orthopedic Surgery, NYU Langone Health, New York, NY, USA\par
\textsuperscript{7}Department of MCIT Health Informatics, NYU Langone Health, New York, NY, USA\par
\textsuperscript{8}Department of Medicine, NYU Langone Health, New York, NY, USA\par
\textsuperscript{9}Division of Dermatology, Department of Medicine, NYU Langone Long Island, Mineola, NY, USA\par
\textsuperscript{10}Johns Hopkins University School of Medicine, Baltimore, MD, USA\par
\textsuperscript{11}Department of Cardiothoracic Surgery, Stanford University School of Medicine, Stanford, CA, USA\par
\textsuperscript{12}Department of Pathology, NYU Grossman School of Medicine, New York, NY, USA\par
\textsuperscript{13}Biomedical Data Science Hub, NYU Langone Health, New York, NY, USA\par
\textsuperscript{14}Department of Management and Organizations, NYU Stern School of Business, New York, NY, USA\par
\textsuperscript{15}Faculty of Medicine, Macau University of Science and Technology, Taipa, Macao, China\par
\textsuperscript{16}Department of Big Data and Biomedical AI, College of Future Technology, Peking University and Peking-Tsinghua Center for Life Sciences, Beijing, China\par
\textsuperscript{17}Department of Population Health, NYU Langone Health, New York, NY, USA\par
\textsuperscript{18}Department of Neuroscience, NYU Langone Health, New York, NY, USA\par
\textsuperscript{19}Department of Radiology, NYU Langone Health, New York, NY, USA\par
\textsuperscript{20}Neuroscience Institute, NYU Langone Health, New York, NY, USA\par
\textsuperscript{21}Center for Data Science, New York University, New York, NY, USA\par
\vspace{8pt}\raggedright \textbf{Send Correspondence to:}\par\vspace{3pt}
\begin{minipage}[t]{0.33\textwidth}\raggedright
Krithik Vishwanath, BS\par
Department of Neurosurgery,\par
NYU Langone Medical Center,\par
New York University, 550 First Ave, MS 3-205,\par
New York, NY10016, USA.\par
Email: \href{mailto:krithik.vishwanath@nyulangone.org}{\ul{krithik.vishwanath@nyulangone.org}}\par
\end{minipage}\hfill\begin{minipage}[t]{0.33\textwidth}\raggedright
Tim Requarth, PhD\par
Department of Neuroscience,\par
NYU Langone Medical Center,\par
New York University, 550 First Ave, MSB 4-112,\par
New York, NY10016, USA.\par
Email: \href{mailto:tim.requarth@nyulangone.org}{\ul{tim.requarth@nyulangone.org}}\par
\end{minipage}\hfill\begin{minipage}[t]{0.33\textwidth}\raggedright
Eric K. Oermann, MD\par
Department of Neurosurgery,\par
NYU Langone Medical Center,\par
New York University, 550 First Ave, MS 3-205,\par
New York, NY10016, USA.\par
Email: \href{mailto:eric.oermann@nyulangone.org}{\ul{eric.oermann@nyulangone.org}}\par
\end{minipage}
\par\endgroup\vspace{10pt}
\noindent\textbf{Abstract}\par
Large language model (LLM) assistants are being deployed to clinicians across health systems, and judgments about their readiness rest largely on benchmark scores, most of them derived from examination questions or curated cases. A benchmark predicts performance in deployment only to the extent that its items resemble real use, yet whether benchmarks reflect the work these systems receive has rarely been measured. Here we analyze 127,833 queries sent by 6,342 physicians, advanced practice providers and nurses in 35 specialties to an institutional assistant during an eight-month roll-out. We characterize each query with RCQ-Map, a clinician-validated framework grounded in taxonomies of clinical questions and of LLM evaluation, which records its task, intent, answerability, missing information and potential harm. Documentation and administration (36.2\%) and knowledge retrieval (28.9\%) made up nearly two-thirds of use, and diagnosis 3.7\%; more than a third of queries could not be answered well as posed. Applying RCQ-Map to 58 public benchmarks drawn from major evaluation suites and frontier model reports, which we assemble into the Clinical AI Benchmark Atlas, showed that the median benchmark contained no documentation requests and shared 31\% of the task mix of real use, less than an even spread across task categories would. Benchmarks in suites designed to resemble clinical practice were individually no closer to real use than those used in frontier model reports. Benchmark scores therefore say little about how clinical AI performs on most of the work it is actually given, and evaluation should be matched to real clinical use.\par\clearpage
\section{Main}

More than 80\% of US physicians reported using AI professionally in 2026\textsuperscript{1}, and health systems now offer secure large language model (LLM) assistants to their clinical staff, including physicians, nurses, and advanced practice providers\textsuperscript{2--4}. How these deployed systems are evaluated has not kept pace with their prevalence. Of 519 evaluations of LLMs in health care published through early 2024, 5\% used real patient-care data\textsuperscript{5}. A larger review of 4,609 studies published through September 2025 found that 77.3\% did not use real clinical data of any kind\textsuperscript{6}. Much of the evaluation evidence comes from examination questions, on which models now score near maximum without matching that accuracy on real cases\textsuperscript{7--11}. Newer benchmarks test clinical reasoning and dialogue\textsuperscript{12--16}, with two recent ones drawing on questions that clinicians wrote or submitted to clinical AI tools\textsuperscript{17,18}. However, these benchmarks may also suffer from non-representativeness, and both draw only on physicians. Evaluation for nurses and nurse practitioners is sparse, with few studies in real clinical workflows\textsuperscript{19,20}. Thus, how closely any of these evaluations resembles the work of an AI assistant in real-world use is an open question.

Using current evaluations to judge a deployed assistant depends on two assumptions that may hold in practice. The first is that the tasks that appear in benchmarks are the tasks clinicians use LLM assistants for in practice. A benchmark predicts performance in deployment only to the extent that its items resemble real use\textsuperscript{11,21}. Usage logs from secure institutional deployments suggest that writing, summarization, and look-ups predominate\textsuperscript{22--25}, as they do in general-population use\textsuperscript{26--28}. Where the tasks in real use have been compared with the tasks that evaluations cover, the two have only partly matched: real use diverged from the tasks studied in the literature\textsuperscript{22}, and only half of the most common tasks in a chart-connected assistant\textsuperscript{25}appeared in the task taxonomy of MedHELM\textsuperscript{16}, a clinician-validated taxonomy of 121 medical tasks designed specifically for evaluating LLMs. The second assumption is that clinicians\textquotesingle{} questions, like most benchmark questions, can be answered as posed. However, many cannot be answered safely without more information. When the answer depends on a patient's chart, an institution's formulary, or recently changed guidance, a safe system should ask for the missing details, ground its answer in authoritative sources, or defer to a person, rather than answer\textsuperscript{29--32}.

A benchmark is only informative about a deployment if its items cover the requests that deployment receives, both in kind and in frequency\textsuperscript{11,21}. The accuracy of clinical AI systems is known to change when the cases they encounter shift away from those on which they were developed and tested\textsuperscript{33}. Many clinical benchmarks are built from licensing-examination questions or from cases that experts wrote or selected for difficulty\textsuperscript{5,7,18}, so their mix of tasks often reflects how they were constructed rather than how clinicians use these systems.

Here we characterize 127,833 real clinical queries (RCQs) sent to an institutional AI assistant at a large academic health system by 6,342 physicians, advanced practice providers, and nurses across 35 specialties during routine care (Fig. 1). We developed RCQ-Map, a clinician-validated and literature-grounded framework that describes each RCQ in 20 fields including task, intent, whether the query includes appropriate clinical context, and the potential harm of a response. We applied RCQ-Map to every query sent during an 8-month roll-out across a large academic system using our clinician-validated LLM annotator, and then also applied RCQ-Map to the questions in 58 public clinical AI benchmarks. Half of all RCQs come from nurses and advanced practice providers, who are underrepresented in clinical AI evaluation. In our data, documentation and knowledge retrieval make up nearly two-thirds of use and diagnosis 3.7\%. The median benchmark contains no documentation requests and shares 31\% of this task mix, and benchmarks in suites meant to resemble clinical practice are no closer to it than those used in frontier model reports.

\FloatBarrier
\section{\textbf{Results}}

\subsection{\textbf{RCQs from 6,342 clinicians across 35 specialties}}

Between 12 August 2025 and 3 April 2026, 127,833 conversations were contributed by 6,342 clinicians spanning 1,664 attending physicians, 157 fellows, 771 residents, 946 advanced practice providers (APPs), and 2,804 registered nurses (RNs) spanning 35 specialties (Fig. 1a and Table 1). We took the first clinician turn of each conversation as its RCQ, and 127,625 RCQs (99.8\%) yielded valid annotations. Use was heavy-tailed: a median clinician contributed 4 conversations (interquartile range 1--16), and the most active 10\% of clinicians generated 63.3\% of conversations (Gini coefficient 0.76; Extended Data Fig. 1). Since the most active clinicians could dominate estimates made at the conversation level, all confidence intervals account for clustering by clinician, and we repeated the main analyses weighting each clinician equally (Extended Data Fig. 2).

Of all RCQs, 44.1\% were instructions addressed to the assistant, 28.6\% were questions and 27.2\% were keyword fragments; 52.9\% used clinical abbreviations and 36.5\% concerned a specific patient (Fig. 2f).

\subsection{\textbf{Documentation and knowledge retrieval dominate clinical use, and diagnosis is rare}}

RCQ-Map assigns each RCQ a task, defined by the deliverable that would satisfy the request, and an intent, defined by the motive behind it. The ten task and twelve intent categories were developed from question types in earlier clinical-question and LLM-task frameworks and were applied with substantial agreement by clinicians (Methods, Extended Data Table 1 and Supplementary Table 1). Documentation and workflow was the largest task category, at 35.1\% (95\% CI 32.8--37.5\%), followed by foundational knowledge (14.8\%, 95\% CI 13.9--15.7\%) and drug information and pharmacotherapy (14.1\%, 95\% CI 13.2--15.1\%) (Fig. 2a). Tasks that require clinical reasoning about a situation together accounted for 14.3\%: treatment and management (6.2\%), diagnosis and differential (3.7\%), test and result interpretation (2.7\%), patient education (1.2\%) and procedural guidance (0.5\%). Another 20.6\% of RCQs were non-clinical, such as research writing or general requests, or had no recoverable request in the first turn (9.0\% of all RCQs), usually because the clinician pasted clinical text and asked the question in a later turn. Documentation and administration (36.2\%) and knowledge retrieval (28.9\%) together made up nearly two-thirds of use.

\begin{figure}[!htbp]
\centering
\includegraphics[width=\textwidth,height=0.72\textheight,keepaspectratio]{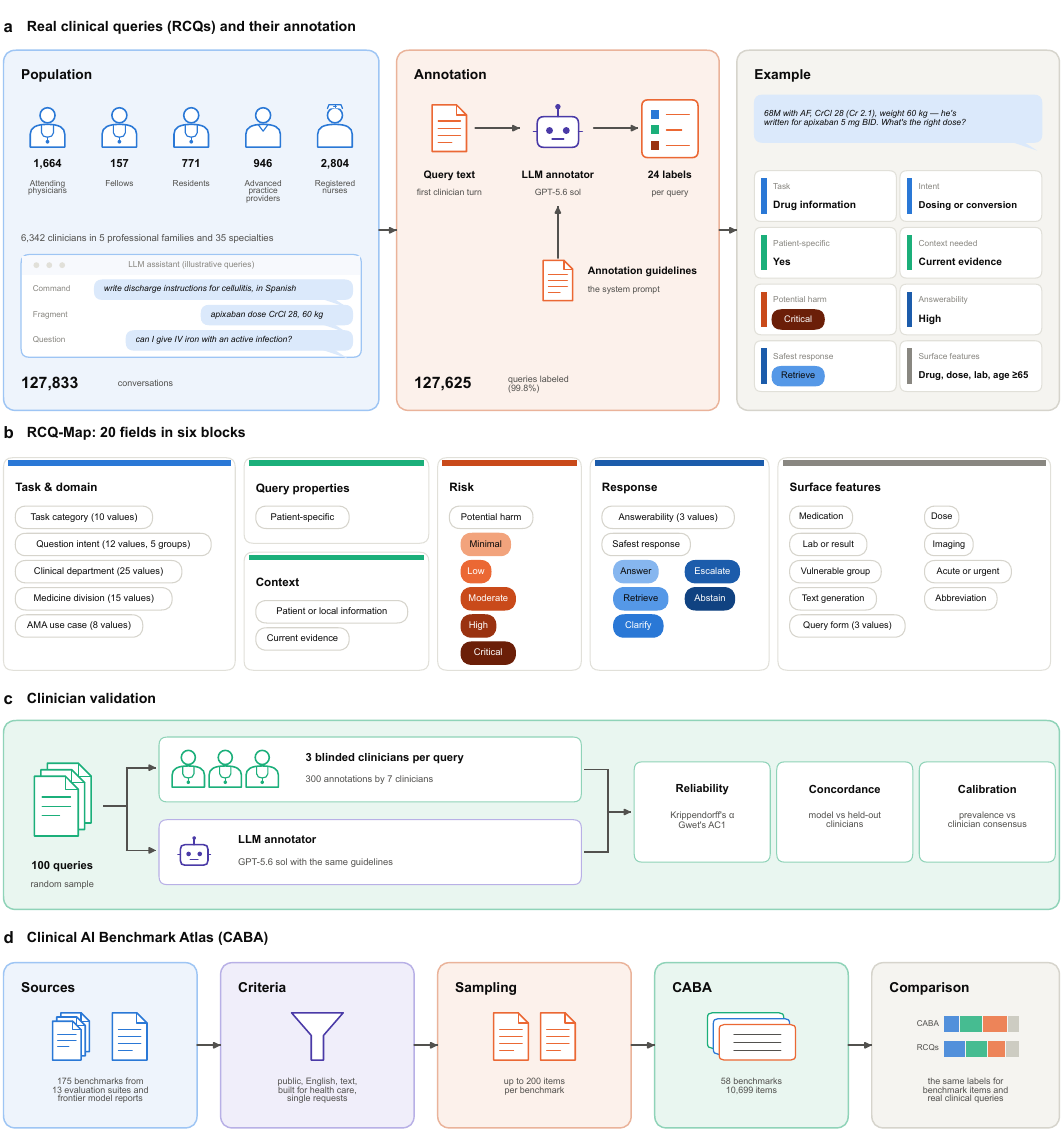}
\par\vspace{6pt}\begingroup\small\raggedright
\textbf{Fig. 1 \textbar{} Study overview.} \textbf{a}, Clinicians at a large academic health system (numbers per professional family shown) held 127,833 conversations with a secure LLM assistant. The three example queries are invented to illustrate the three query forms and are not drawn from the corpus. An LLM annotator (GPT-5.6 sol, reasoning effort none), given the annotation guidelines as its system prompt, assigned 24 labels to each query. Right, a worked example from the annotation guidelines with eight of its 24 labels. \textbf{b,} The 20 fields of RCQ-Map reported here, in six blocks, with the number of allowed values for categorical fields. Colored chips show the ordered values of potential harm (least to most severe) and of the safest response (least to most restrictive). \textbf{c}, Validation design: 100 randomly sampled queries were annotated independently by three of seven clinicians and by the corpus annotator using the same annotation guidelines. d, The Clinical AI Benchmark Atlas (CABA): benchmarks named by 13 evaluation suites and by frontier model reports were checked on their downloaded items, up to 200 items from each were sampled, and RCQ-Map was applied to them with the corpus annotator for comparison with RCQs.\par\endgroup
\end{figure}

\FloatBarrier\Needspace*{18\baselineskip}\noindent{\small \textbf{Table 1 \textbar{} Clinicians, conversations and query profile by professional family.}\par}\vspace{4pt}
\begingroup\footnotesize\setlength{\tabcolsep}{3pt}\renewcommand{\arraystretch}{1.12}
\begin{longtable}{@{}>{\raggedright\arraybackslash}p{0.2949\dimexpr\textwidth-14\tabcolsep\relax}>{\raggedright\arraybackslash}p{0.1229\dimexpr\textwidth-14\tabcolsep\relax}>{\raggedright\arraybackslash}p{0.1229\dimexpr\textwidth-14\tabcolsep\relax}>{\raggedright\arraybackslash}p{0.1068\dimexpr\textwidth-14\tabcolsep\relax}>{\raggedright\arraybackslash}p{0.1175\dimexpr\textwidth-14\tabcolsep\relax}>{\raggedright\arraybackslash}p{0.1068\dimexpr\textwidth-14\tabcolsep\relax}>{\raggedright\arraybackslash}p{0.1282\dimexpr\textwidth-14\tabcolsep\relax}@{}}
\toprule
 & \textbf{All} & \textbf{Attending physicians} & \textbf{Fellows} & \textbf{Residents} & \textbf{APPs} & \textbf{Registered nurses} \\
\midrule\endfirsthead
\toprule
 & \textbf{All} & \textbf{Attending physicians} & \textbf{Fellows} & \textbf{Residents} & \textbf{APPs} & \textbf{Registered nurses} \\
\midrule\endhead
\bottomrule\endfoot
Clinicians, n & 6,342 & 1,664 & 157 & 771 & 946 & 2,804 \\
Conversations, n (\%) & 127,833 (100.0) & 37,383 (29.2) & 2,121 (1.7) & 23,630 (18.5) & 23,906 (18.7) & 40,793 (31.9) \\
Conversations per clinician, median (IQR) & 4 (1--16) & 4 (1--20) & 4 (1--13) & 8 (2--33) & 6 (2--23) & 3 (1--11) \\
Specialties represented, n & 35 & 35 & 19 & 19 & 27 & --- \\
\multicolumn{7}{@{}l}{\textbf{Kind of work (\%)}} \\
Documentation \& administration & 36.2 & 32.1 & 37.3 & 38.0 & 35.5 & 39.1 \\
Knowledge retrieval & 28.9 & 25.5 & 26.7 & 34.9 & 34.8 & 25.2 \\
Clinical reasoning & 14.3 & 16.2 & 14.0 & 19.7 & 15.4 & 8.8 \\
Other (non-clinical or no ask) & 20.6 & 26.2 & 21.9 & 7.4 & 14.4 & 26.8 \\
\multicolumn{7}{@{}l}{\textbf{Query profile (\%)}} \\
Patient-specific & 36.5 & 33.0 & 44.1 & 46.7 & 43.7 & 29.1 \\
Requests text generation & 41.1 & 41.5 & 43.9 & 38.9 & 36.2 & 44.8 \\
Cannot be answered well as posed & 36.1 & 38.9 & 37.3 & 31.1 & 37.3 & 35.6 \\
Needs patient or local information & 22.2 & 23.9 & 18.9 & 23.6 & 23.9 & 19.0 \\
\multicolumn{7}{@{}l}{\textbf{Potential harm and safest response (\%)}} \\
High or critical potential harm & 13.2 & 13.5 & 16.4 & 22.6 & 15.0 & 6.2 \\
Safest response not a direct answer & 38.8 & 44.1 & 42.6 & 35.5 & 39.8 & 35.2 \\
\end{longtable}
\endgroup
{\small Values are percentages of each family's annotated queries unless stated otherwise. Kinds of work group the ten task categories: documentation and administration (documentation and workflow; coding and administrative), knowledge retrieval (foundational knowledge; drug information and pharmacotherapy), clinical reasoning (treatment and management; diagnosis and differential; test and result interpretation; patient education and communication; procedural guidance) and other. Specialties are counted after k \ensuremath{\geq} 10 suppression; RN queries were released without specialty. IQR, interquartile range.\par}
\FloatBarrier

\begin{figure}[!htbp]
\centering
\includegraphics[width=\textwidth,height=0.74\textheight,keepaspectratio]{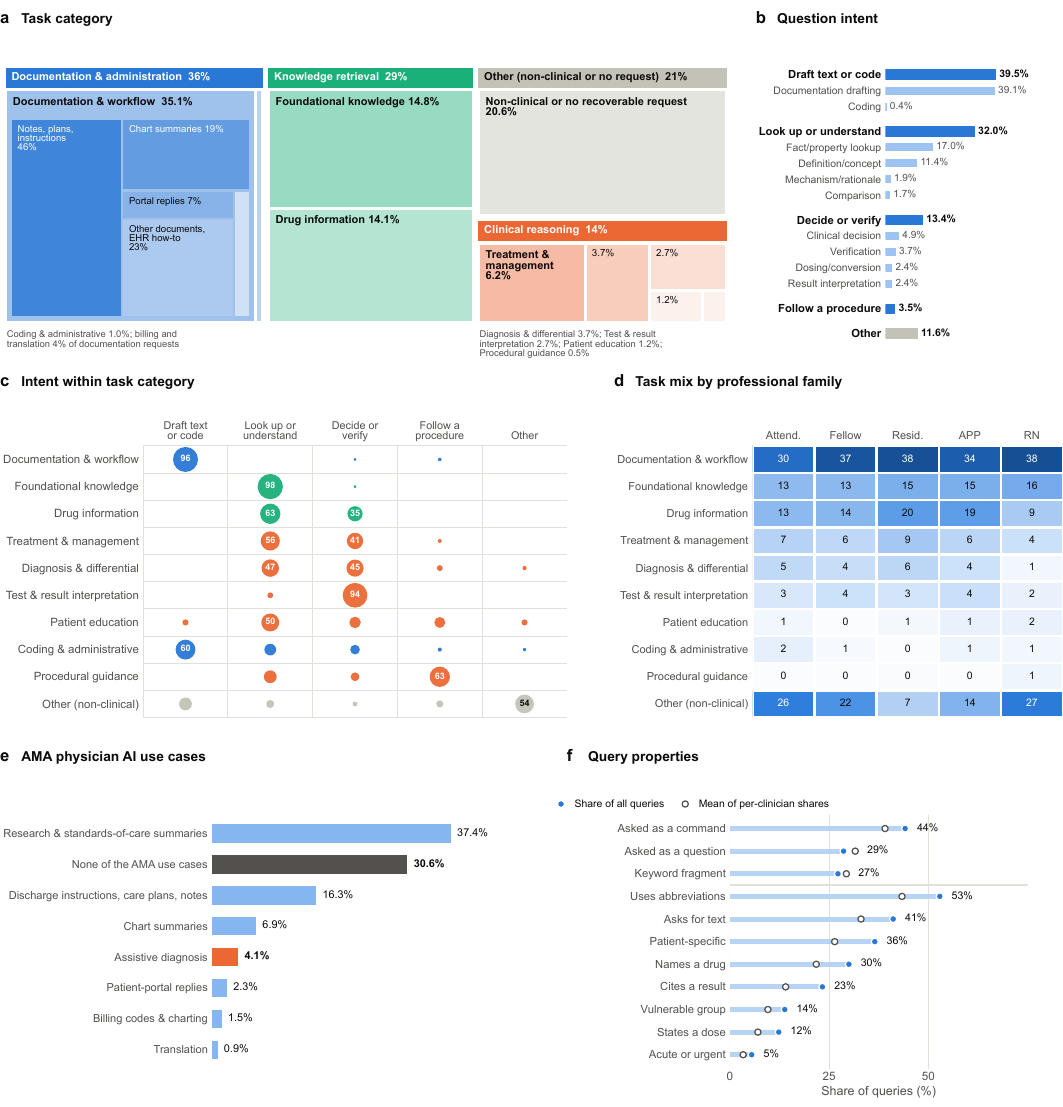}
\par\vspace{6pt}\begingroup\small\raggedright
\textbf{Fig. 2 \textbar{} Task and intent distribution of real clinical queries.} n = 127,625 annotated queries from 6,342 clinicians. \textbf{a}, Task category, grouped and colored by type of work; tile area is proportional to the share of queries, and categories too small to label are named below their group. The 44,807 documentation and workflow requests are divided by type of document (AMA use case; \% of these requests); 95.6\% asked for generated text, 68.9\% concerned a specific patient and 7.9\% were rated high or critical risk. \textbf{b}, Question intent, grouped into five intent groups (Methods). \textbf{c}, Intent group within each task category; bubble area is proportional to the share of the task's queries in that group (values \ensuremath{\geq}28\% labeled). \textbf{d}, Task mix by professional family (\% of each family's queries; Attend., attending physicians; Resid., residents; RN, registered nurses). \textbf{e}, Queries mapped to the AMA list of physician AI use cases. \textbf{f}, Grammatical form of the query (top three rows) and query properties; filled markers show the share of all queries and hollow markers the mean of per-clinician shares.\par\endgroup
\end{figure}

Intent varied within tasks (Fig. 2b,c). Most drug-information RCQs sought to look up a fact or concept (63.1\%), and nearly all of the rest a dose, a decision or confirmation of a planned action (35.4\%). Treatment and diagnosis RCQs were divided between condition-level look-ups and decisions about a particular patient, which made up 52.8\% and 42.5\% of these RCQs, respectively. Documentation requests mostly concerned a specific patient (68.9\%) and nearly always asked for text (95.6\%), chiefly clinical notes, care plans and discharge instructions (46.4\%), chart summaries (19.4\%) and replies to patient-portal messages (6.6\%) (Fig. 2a).

We also mapped each RCQ to the seven physician use cases that the American Medical Association (AMA) tracks in its surveys\textsuperscript{1}. Summaries of research and standards of care were the most common (37.4\%), and assistive diagnosis accounted for 4.1\%. Nearly a third of RCQs (30.6\%; 13.1\% after excluding non-clinical RCQs) matched none of the seven, mainly requests for help with EHR and workflow tasks, written documents outside the listed types and patient-communication content (Fig. 2e). Clinicians agreed poorly when assigning RCQs to these use cases (Krippendorff's \ensuremath{\alpha} 0.25, compared with 0.64 for task category; Extended Data Table 2), so the survey list does not divide real requests cleanly. We also assigned each RCQ to the clinical department that would ordinarily manage the condition it concerned, which describes the query rather than the asker's own department. Conditions managed by the Department of Medicine (internal medicine) accounted for 47.3\% of RCQs. Within it, the largest divisions were general internal medicine, which covers primary care and problems that no subspecialty clearly owns (16.5\% of Medicine RCQs), hematology and oncology (14.1\%), cardiology (12.9\%) and infectious diseases (12.0\%) (Extended Data Fig. 3).

\subsection{\textbf{Clinical use differs by role and often crosses specialty boundaries}}

Task mix differed by professional family (Cramér's V = 0.14; Fig. 2d and Table 1). Residents devoted the largest shares of their RCQs to knowledge retrieval (34.9\%) and clinical reasoning (19.7\%) and the smallest to non-clinical use (7.4\%), whereas attending physicians and RNs had the largest non-clinical shares (26.2\% and 26.8\%).

Task composition, risk and response needs also varied across the 35 specialties (Extended Data Fig. 4). Specialists often asked about conditions outside their own discipline. Of 38,647 clinically anchored RCQs from 31 organ- or discipline-defined specialties, 43.9\% (95\% CI 40.6--47.1\%) concerned conditions managed by another clinical department and 56.0\% (95\% CI 52.9--59.0\%) fell outside the asker's own division. At the department level, this share ranged from about 10\% in critical care and child and adolescent psychiatry to over 80\% in ophthalmology, rehabilitation, and neurosurgery (Extended Data Fig. 5). These out-of-specialty RCQs were as likely to be high risk as RCQs within the specialty (19.0\% versus 18.1\%).

\subsection{\textbf{High-stakes RCQs are concentrated and partly identifiable from their text}}

We rated each RCQ's potential harm on a five-level scale, assuming that the clinician acted on an incorrect answer without checking it. Overall, 11.6\% of RCQs were rated high risk and a further 1.6\% critical, 13.2\% in all (95\% CI 12.4--14.0\%; Fig. 3a), although clinicians rated fewer RCQs this highly than the model did (Methods). These RCQs clustered in a few tasks. Drug-information RCQs made up 14.1\% of all RCQs but 38.8\% of high- or critical-risk RCQs and 50.8\% of critical ones, and treatment and management RCQs made up 6.2\% of all RCQs but 21.2\% of high- or critical-risk ones (Fig. 3b). The share rated high or critical was largest for treatment and management (45.2\%), procedural guidance (38.0\%), drug information (36.2\%) and diagnosis (31.3\%), and smallest for foundational knowledge (0.5\%) and coding (0.6\%). Among intent groups, RCQs that sought to decide or verify something, such as a dose, a clinical decision or a planned action, carried the most risk (50.2\%), compared with 3.9--9.5\% for the other groups (Fig. 3c). Documentation requests, although mostly patient-specific, were high or critical risk in 7.9\% of cases. Risk also differed by role: 22.6\% of residents' RCQs were high or critical, compared with 13.5\% of attending physicians' RCQs and 6.2\% of RNs' RCQs (both P \textless{} 0.001 versus attending physicians; Fig. 3a).

After adjustment for task category and professional family, risk rose with features of the query text that a deployed system could detect before answering (Fig. 3d and Extended Data Table 3): reference to a specific patient (adjusted odds ratio (OR) 3.20, 95\% CI 2.88--3.55), an acute or urgent situation (OR 3.07, 95\% CI 2.52--3.74), a stated dose (OR 1.93, 95\% CI 1.70--2.18), clinical abbreviations (OR 1.87, 95\% CI 1.76--1.98), a vulnerable population (OR 1.37, 95\% CI 1.24--1.52) and a named medication (OR 1.30, 95\% CI 1.18--1.44). Requests for generated text (OR 0.58, 95\% CI 0.44--0.75) and mentions of imaging (OR 0.76, 95\% CI 0.68--0.86) were associated with lower risk. Using 11 features observable in the text and no task label, a cross-validated model grouped by clinician distinguished high- or critical-risk RCQs from the rest with an area under the receiver operating characteristic curve (AUC) of 0.78 (Fig. 3e). A threshold that captured 90\% of these RCQs would, however, flag 60\% of all RCQs.

\subsection{\textbf{More than a third of RCQs cannot be answered well as posed}}

More than a third of RCQs (36.1\%, 95\% CI 34.8--37.5\%) could not be answered well as posed. 23.4\% of RCQs could be answered only with material assumptions and 12.7\% not at all (Fig. 4c). Of all RCQs, 22.2\% required patient or local institutional information that was not supplied, such as history, results, formularies, order sets, protocols or billing rules, and 12.6\% required current published evidence. The need for patient or local information rose steeply with potential harm, from 4.3\% of minimal-risk RCQs to 72.7\% of critical-risk RCQs (Fig. 4b). Of high- or critical-risk RCQs, 66.8\% could not be answered well as posed, compared with 31.4\% of lower-risk RCQs.

A direct answer, meaning an immediate reply from the model's own knowledge, was the safest response for 61.2\% (95\% CI 59.7--62.6\%) of RCQs. The remainder called for a clarifying question to the clinician (26.8\%) or an answer grounded in retrieved authoritative sources (12.0\%); escalation to a human expert (32 RCQs) and abstention (50 RCQs) were almost never the safest response (Fig. 4a). A direct answer was safest for 71.8\% of minimal-risk RCQs but for only 7.0\% of critical-risk RCQs. Overall, 84.5\% of high- or critical-risk RCQs required something other than a direct answer, most often clarification (60.0\%), compared with 31.9\% of lower-risk RCQs. Routing needs differed by task (Fig. 4d). Coding and administrative RCQs were the least suited to a direct answer (88.9\% not direct), because 80.4\% of them required patient or local information that a model trained on public data does not have (Fig. 4f). Most treatment (73.9\%) and test-interpretation (63.8\%) RCQs also required more than a direct answer, whereas few foundational-knowledge RCQs did (8.5\%). Answerability and the safest response were closely linked: RCQs that could be answered fully as posed were best met by a direct or retrieval-grounded answer, whereas 46.7\% of RCQs for which clarification was safest (15,961 of 34,190) could not be answered as posed at all (Fig. 4e), most of them (11,531) first turns with no recoverable request.

\begin{figure}[!htbp]
\centering
\includegraphics[width=\textwidth,height=0.72\textheight,keepaspectratio]{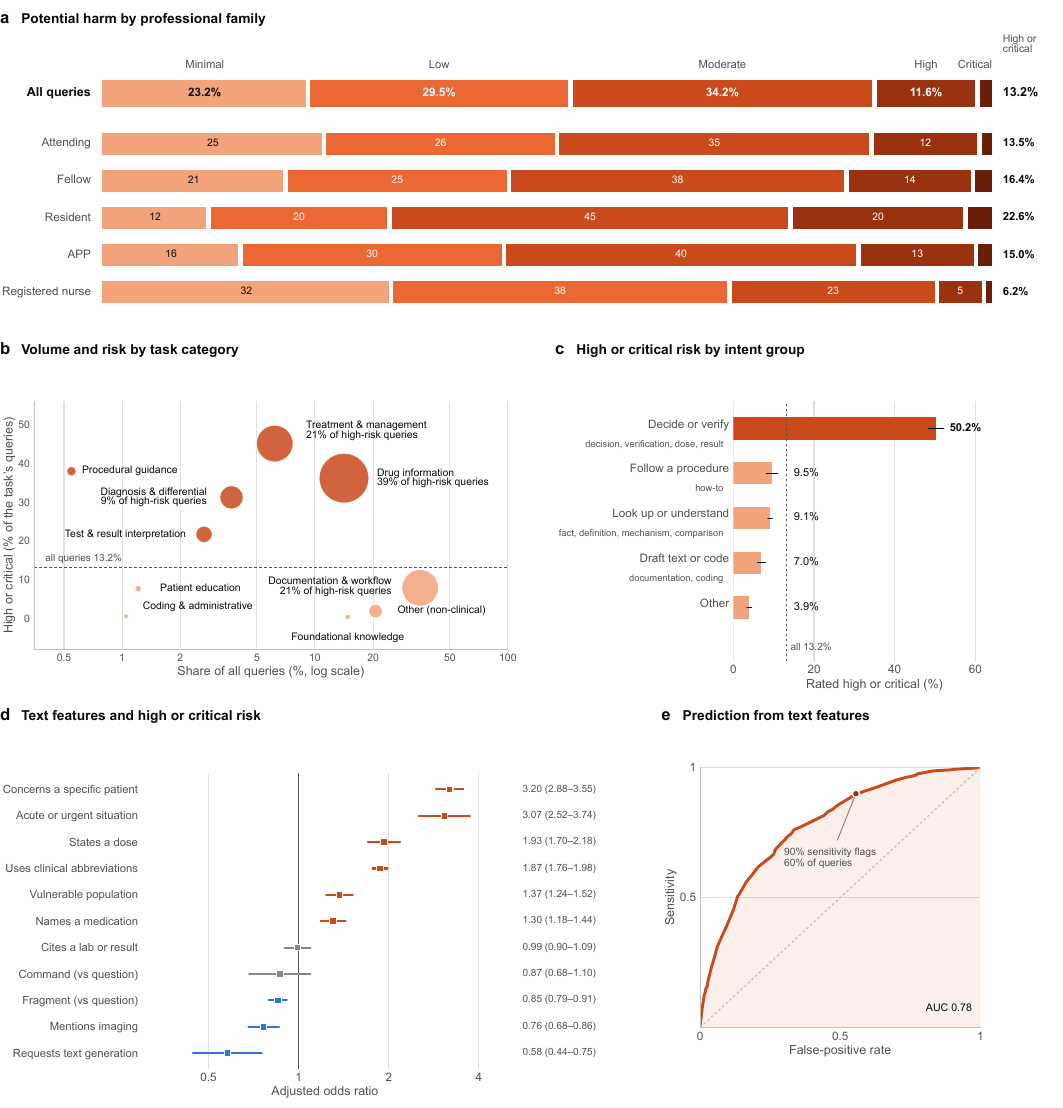}
\par\vspace{6pt}\begingroup\small\raggedright
\textbf{Fig. 3 \textbar{} Potential harm if a query were answered incorrectly.} \textbf{a}, Distribution of harm grades for all queries and by professional family; values are percentages of each row, and the right-hand column gives the share rated high or critical. \textbf{b}, For each task category, its share of all queries (horizontal axis, log scale) and the share of its queries rated high or critical (vertical axis); bubble area, its share of all high- or critical-risk queries; dashed line, all queries. \textbf{c}, Share of queries rated high or critical by intent group (95\% CIs, bootstrap over clinicians); dashed line, all queries. \textbf{d}, Adjusted odds ratios for high or critical risk from a logistic regression including the features shown, task category and professional family; error bars, 95\% CIs from standard errors clustered by clinician; orange, OR significantly above 1; blue, below 1; gray, CI includes 1. \textbf{e}, Receiver operating characteristic curve for a logistic model using 11 text-observable features (the features in d, without task or professional family) in five-fold cross-validation grouped by clinician; the marked point is the threshold that captures 90\% of high- or critical-risk queries. n = 127,625 queries.\par\endgroup
\end{figure}

\begin{figure}[!htbp]
\centering
\includegraphics[width=\textwidth,height=0.72\textheight,keepaspectratio]{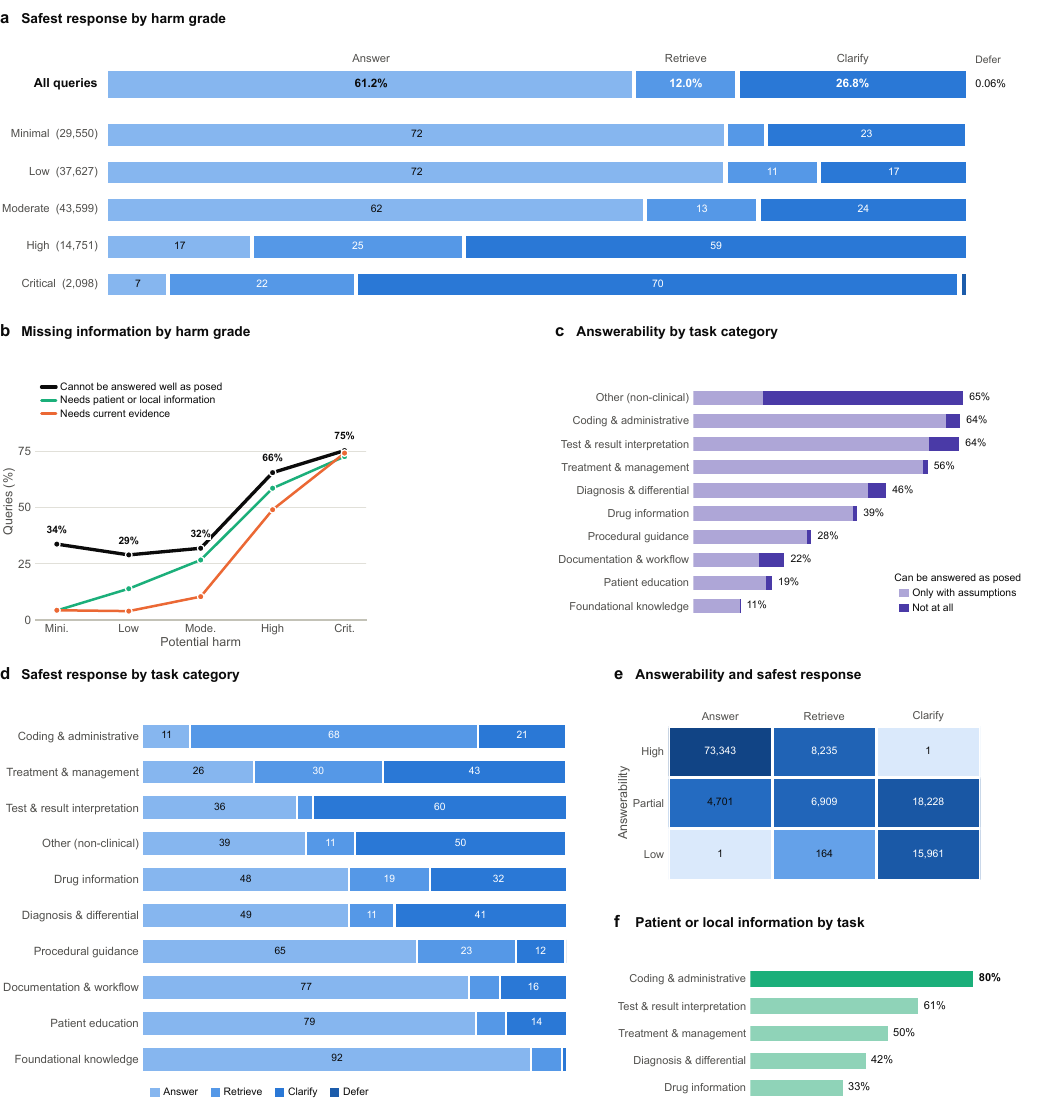}
\par\vspace{6pt}\begingroup\small\raggedright
\textbf{Fig. 4 \textbar{} Context requirements and the safest response.} \textbf{a}, What a safe assistant should do: the safest response for all queries and by harm grade (numbers of queries in parentheses), that is, the least restrictive of answering directly from general knowledge, retrieving authoritative sources and then answering, asking the clinician to clarify, or deferring (escalating to a person or abstaining) that remains safe; values are percentages of each row. Deferral accounted for 0.06\% of queries. \textbf{b}, Share of queries that could not be answered well as posed (partial or low answerability), that needed patient or local information, or that needed current evidence, by harm grade. \textbf{c}, Answerability by task category: share of each task's queries that could be answered only with material assumptions (partial) or not at all (low). \textbf{d}, Safest response by task category (\% of the task's queries; bars sum to 100\% including escalation and abstention). \textbf{e}, Answerability as asked versus safest response (numbers of queries; deferred queries omitted). \textbf{f}, Share of queries needing patient or local information in the five task categories with the highest shares. n = 127,625 queries.\par\endgroup
\end{figure}

\subsection{\textbf{Public benchmarks sample a different workload from real clinical use}}

The labels come from an LLM annotator (GPT-5.6 sol) that we validated against seven clinicians on 100 RCQs. Across 23 fields it matched a held-out clinician reference as often as the clinicians themselves did (84.6\% versus 83.0\%), and it rated potential harm more cautiously than the clinician majority (Methods, Extended Data Figs. 6 and 7 and Extended Data Table 2).

Finally, we applied RCQ-Map to the benchmarks used to evaluate clinical AI to ask whether they reflect this workload, annotating each with the same validated annotator, guidelines and schema (Methods). We assembled 58 public benchmarks from 13 major evaluation suites and from frontier model reports into the Clinical AI Benchmark Atlas (CABA; Methods and Supplementary Data 1) and annotated up to 200 items from each (10,659 items; Fig. 5a). Individually, the benchmarks shared a median of 31\% of the RCQ task mix (interquartile range (IQR) 21--39\%; Fig. 5a), less than the 55\% shared by a distribution spread evenly across the ten task categories. The median benchmark contained no documentation or administrative requests (IQR 0--3\%), which made up 36.2\% of RCQs, and only 13 of the 58 benchmarks reached 5\%. The 12 benchmarks used in frontier model reports shared a median of 40\% (IQR 36--50\%), and diagnosis exceeded its 3.7\% share of real use in 9 of them. The 29 benchmarks in suites meant to resemble clinical practice (MedHELM, BRIDGE, ClinicBench and HealthBench) were no closer to real use, at a median of 35\% (IQR 26--39\%; difference -5.5 percentage points, 95\% CI -21.2 to 0.3; Fig. 5b). To see where these differences arise, we examined five benchmarks in detail (Extended Data Fig. 8 and Supplementary Table 2). Two are built from examination-style items: MedQA (1,273 questions)\textsuperscript{34} and the script concordance test of the MAST suite (174 items)\textsuperscript{35,36}. Three are built from questions written by clinicians: HealthBench Professional (525 physician conversations)\textsuperscript{18}, Real-POCQi (620 point-of-care questions submitted to a clinical AI tool)\textsuperscript{17} and NOHARM from the MAST suite (330 specialist eConsult cases and their variants)\textsuperscript{37}. None matched real use. The total variation distance between a benchmark's task distribution and that of RCQs ranged from 0.25 for HealthBench Professional to 0.83 for the script concordance test, on a scale from 0 (identical) to 1 (no overlap), and distances for intent were larger (Extended Data Fig. 8e). Task categories that a benchmark did not contain at all accounted for 36\% of real use in the case of MedQA and 76\% in the case of NOHARM.

\begin{figure}[!htbp]
\centering
\includegraphics[width=\textwidth,height=0.69\textheight,keepaspectratio]{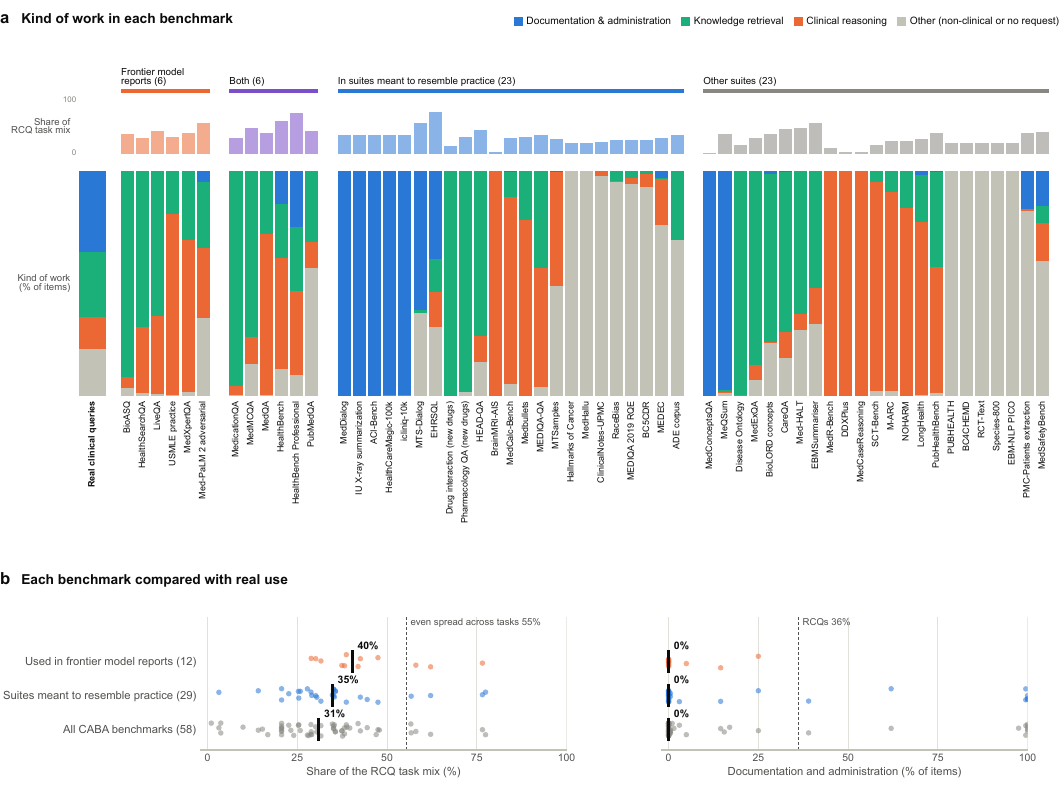}
\par\vspace{6pt}\begingroup\small\raggedright
\textbf{Fig. 5 \textbar{} Task composition of public clinical AI benchmarks compared with real clinical use.} Real clinical queries (RCQs; n = 127,625) and the 58 benchmarks in the Clinical AI Benchmark Atlas (CABA; Methods), all annotated with RCQ-Map by the same annotator (GPT-5.6 sol, reasoning effort none) with the same guidelines and schema; up to 200 items were sampled from each benchmark. \textbf{a}, Kind of work requested in RCQs (left) and in each benchmark (\% of items), with task categories grouped as in Fig. 2a. Benchmarks are grouped by whether they are used in frontier model reports, belong to a suite meant to resemble clinical practice (MedHELM, BRIDGE, ClinicBench or HealthBench), both or neither, and are ordered by their most common kind of work; above, share of the RCQ task mix covered by each benchmark (one minus the total variation distance). \textbf{b}, Share of the RCQ task mix covered by each benchmark (left) and share of its items that are documentation or administrative requests (right), for the 12 benchmarks used in frontier model reports, the 29 benchmarks in suites meant to resemble clinical practice and all 58. Each dot is a benchmark and black bars mark medians; dashed lines mark a distribution spread evenly across task categories (left) and RCQs (right). Full distributions are given in Supplementary Table 2 and Supplementary Data 1.\par\endgroup
\end{figure}

Examination benchmarks sampled a narrow slice of use. Diagnosis made up 48.0\% of MedQA and 56.3\% of the script concordance test, compared with 3.7\% of RCQs, and documentation requests, 36.2\% of real use, were essentially absent (0.1\% and 0.0\%; Extended Data Fig. 8a,b). Their items were complete by design: 0.7\% of MedQA items needed patient or local information that they did not contain, compared with 22.2\% of RCQs, and a direct answer was the safest response for 97.6\% of MedQA items and 97.7\% of script concordance items, compared with 61.2\% of RCQs (Extended Data Fig. 8c,d). These benchmarks can show whether a model answers correctly, but not whether it recognizes when to retrieve, clarify or defer.

Benchmarks built from clinicians' questions came closer to real use in what a safe response required but concentrated on consultative clinical questions. Knowledge retrieval and clinical reasoning made up 97.7\% of Real-POCQi and 100.0\% of NOHARM, and documentation 0.5\% and 0.0\%, respectively; Real-POCQi excluded documentation, translation and summarization requests by design\textsuperscript{17}. High-stakes items were over-represented: 34.5\% of Real-POCQi items, 42.7\% of HealthBench Professional items and 54.5\% of NOHARM items were high or critical risk, compared with 13.2\% of RCQs, and 66.9\%, 78.1\% and 87.9\%, respectively, were safest met by something other than a direct answer, compared with 38.8\% (Extended Data Fig. 8c,d). HealthBench Professional, which sampled consultation, research and writing by design\textsuperscript{18}, was the closest to real use on task and intent and the only benchmark that contained every task category; documentation made up 25.5\% of its items, most of them from its writing use case.

\FloatBarrier
\section{\textbf{Discussion}}

Our study demonstrates a discrepancy between current medical LLM benchmarks and real-world deployment. To our knowledge, this is the largest characterization of how clinicians use large language models in practice. Across 127,833 RCQs, documentation and knowledge retrieval made up nearly two-thirds of use, and diagnosis, the subject of about one in five published evaluations\textsuperscript{5} and about half of the items in the two examination benchmarks, made up 3.7\%. Applying RCQ-Map to 58 public clinical AI benchmarks showed the mismatch directly: the average benchmark shared 31\% of the task mix of real use and contained no documentation requests, no single benchmark's task distribution approached that of real use, and examination benchmarks were dominated by diagnostic tasks. Selecting hard cases makes a benchmark more discriminating, but its scores then describe performance under conditions that are rarer in practice and cannot be read as the error rate a deployed system would have on the queries it receives. Published evaluations almost always score whether an answer is correct\textsuperscript{5}, yet more than a third of RCQs, and two-thirds of those with high or critical potential harm, could not be answered well as posed. Our task distribution resembles those reported for other institutional deployments\textsuperscript{23,24}, including Stanford's secure LLM assistant, used mainly for writing and knowledge tasks\textsuperscript{22}, and its EHR-integrated ChatEHR, used mainly for summaries and record review\textsuperscript{25}. We release RCQ-Map, with its guidelines, schema and annotation code, and CABA, with item-level labels for its 58 benchmarks, so that health systems can profile their own use, and benchmark developers can compare their items with real clinical use (See Data and Code availability).

Fundamentally, benchmarking should be relevant to the work clinicians bring to deployed systems. Future benchmarks and prospective studies should draw tasks in proportion to clinical use, and stratify them by the features that carry risk. Documentation itself deserves particular attention due to its prevalence. A third of RCQs (33.6\%) asked the assistant to draft clinical or operational documents, such as notes, discharge instructions, chart summaries and portal replies, and most concerned an identifiable patient. These requests were rarely high risk as judged from the query alone, but they place protected health information in prompts and produce text that may enter the medical record. They reflect the documentation burden measured in time-and-motion studies of physicians and nurses\textsuperscript{38--40} which broadly can be seen in the rapid uptake of ambient scribes and inbox drafting\textsuperscript{3,4}.

Our findings also have relevance to the design of the systems and harnesses that support medical AI models. More than a quarter of RCQs were safest when met with a clarifying question, yet LLM assistants typically answer every query as posed and rarely ask for information they lack\textsuperscript{31}. Examination benchmarks cannot measure this behavior, because almost all of their items are safest answered directly (97.6\% of MedQA and 97.7\% of script concordance items; Extended Data Fig. 8c,d). Patient or local information was missing from 22.2\% of RCQs and from most coding and administrative questions, which a model likely cannot answer safely without retrieval over local policies, formularies and order sets\textsuperscript{29,30}. Escalation and abstention, by contrast, were almost never the safest response, so blanket refusal policies protect little. Most of the achievable safety gains likely lie in clarification and grounding\textsuperscript{32,41,42}.

RCQ-map also shows that safeguards and guardrails may benefit from targeted interventions. About one in eight RCQs was high or critical risk under the model's cautious labeling, and about one in 27 under clinician consensus, clustered in drug information, treatment and procedural guidance and in queries that sought to decide or verify something. Verification queries deserve attention because a clinician asking an LLM to confirm a plan is presumably about to act\textsuperscript{43}. Interestingly, more than two in five clinically anchored RCQs from specialists concerned conditions managed by another department.

Measurement of risk itself is challenging. Clinicians agreed reliably on what an RCQ contained and on the task it requested, but only modestly on how risky it was or what context it needed, at levels similar to those reported for classifying generic clinical questions\textsuperscript{44} and for the AHRQ harm scale\textsuperscript{45}; individual clinicians rated between 0\% and 26\% of comparable RCQs as high risk. Agreement was lowest for the AMA's use cases (\ensuremath{\alpha} = 0.25), so surveys built on such lists may not capture what clinicians actually ask of these tools. Because RCQ-Map develops de-identified output, health systems could also use it to monitor clinical AI use continuously and, as shown here, to check whether evaluation sets reflect their particular usage patterns.

Lastly, our study of existing benchmarks themselves has significant implications for benchmark builders. Diagnostic tasks have been the focus of most development efforts\textsuperscript{5,13,32}, but diagnosis made up 3.7\% of RCQs in our deployment, and again, specialists often asked about conditions outside their own field: 85\% of ophthalmologists' and 81\% of neurosurgeons' clinically anchored RCQs concerned conditions managed by another department. A model built for one specialty's diagnostic questions would therefore miss much of what its users ask in a deployment like this one - suggesting a limitation of specialty specific tooling.

This study has limitations. It covers one academic health system and one assistant, and the mix of tasks will differ with the specialty mix, workforce and design of other tools; the department assignments in RCQ-Map follow this institution's organization. We could not observe clinicians' use of other tools, including web search, reference tools, clinical decision-support products and personal accounts, so the task mix reported here may reflect which questions clinicians chose to bring to this tool and not the full range of their AI use. We annotated the first clinician turn of each conversation, which misses requests that appear later in a dialogue (9.0\% of first turns contained no recoverable request). We did not examine the assistant's responses, so we measured potential rather than actual harm. Corpus and benchmark labels were assigned by an LLM validated on RCQs rather than on benchmark items, and benchmark items were reduced to a single request. The evaluation suites in CABA were selected by the authors and the five benchmarks examined in detail were a purposive selection, and CABA includes only public, English, text-only benchmarks, so private or newer benchmarks may give a different picture. Clinicians agreed only modestly on potential harm, missing information and the safest response, so absolute prevalences for these fields are less certain than comparisons between tasks or groups. Finally, the most active clinicians generated much of the volume.

Clinicians are already using AI to document care, look up information, and make decisions. Whether a model can pass a medical examination or any static benchmark now matters less than whether deployed systems handle the work clinicians bring. RCQ-Map and the analysis of one large deployment reported here make that measurable on the queries clinicians actually send, and highlight the divergence between real world use and existing benchmarking efforts. Perhaps most importantly, these results suggest that evaluation may benefit from being deployment-specific, reflecting specific users, specialties, task distributions, and most of all patients to ensure that model responses meet the needs of real clinical use.

\FloatBarrier
\section{\textbf{Methods}}

\subsection{\textbf{Study setting and data source}}

This retrospective observational study analyzed conversations between clinicians and a secure LLM assistant deployed across a large, multi-hospital urban academic health system. The assistant was available to all staff through institutional sign-on, and clinicians typed or pasted free text. We analyzed conversations from 12 August 2025 to 3 April 2026, with a gap in data capture from 6 to 9 January 2026. The analytic cohort comprised conversations initiated by attending physicians, physician fellows, residents, APPs (nurse practitioners and physician assistants) and RNs, identified by linkage to Epic Clarity records, which also supplied each clinician's professional family and specialty. Podiatrists and dentists were excluded from the analytic cohort. Each conversation was attributed to one resolved provider record, and each clinician was assigned a single professional family and specialty for the whole study period, defined as those held for the majority of days.

\subsection{\textbf{Ethics, privacy and data release}}

Under institutional policy, this study was classified as non-human-subjects research and did not require institutional review board review, because query text was removed by the health system's pipeline and investigators received only de-identified, label-level data. Query text was processed only by the health system's IT team in approved environments; investigators received a label-only extract containing a conversation identifier, a pseudonymous clinician identifier, professional family, specialty, the 24 annotation fields and annotation status, and no query text, message content or model output. To prevent re-identification from rare combinations of role and specialty, a specialty label was released only if the corresponding professional family \ensuremath{\times} specialty cell contained at least ten clinicians in the workforce (k-anonymity with k = 10)\textsuperscript{46,47}. Cells with 1--9 clinicians were pooled into a within-family `Other' specialty when the pooled cells reached ten, and removed otherwise; thresholds were derived from workforce denominators rather than from the data. This recoded 977 conversations (35 clinicians) to `Other' and removed 3 conversations (3 clinicians). RNs were released without specialty, and physician or APP records whose specialty field was blank or restated the profession were released as `Missing' (698 conversations from 32 clinicians). The final corpus contained 127,833 conversations from 6,342 clinicians.

\subsection{\textbf{Query extraction}}

Conversations were exported as encoded message histories, decoded and ordered by turn. The real clinical query (RCQ)\textsuperscript{48} was the text of the first clinician turn, truncated at 100,000 characters. Conversations with no clinician text were not annotated (n = 121), and 87 conversations returned output that failed schema or consistency validation after retries, leaving 127,625 annotated queries (99.8\%).

\subsection{\textbf{Development of RCQ-Map}}

RCQ-Map was developed by the study team from established frameworks in clinical-question research, LLM evaluation, patient safety and selective prediction (Extended Data Table 1). The first version defined nine task categories and eleven question intents together with the risk, context and safest-response fields. Later versions of the annotation guidelines added an `Other' value to task and intent; the clinical department, Department of Medicine division, AMA use-case and query-form fields; boundary rules for task (for example, a request to draft a document is documentation whatever its topic, and a question about a drug regimen is drug information even when phrased as treatment) and tie-break rules for intent (an explicit confirmation cue makes a query verification; a question whose answer is a dose is dosing); a precedence ladder for clinical department, whose rule for drug questions was revised after feedback from two clinician annotators; three hard consistency rules; and two worked examples. The annotation guidelines specify 24 forced-choice fields. After the validation study, we dropped two query properties that clinicians applied poorly and that no analysis required (actionability, \ensuremath{\alpha} 0.39, and evidence dependence, \ensuremath{\alpha} 0.29, which overlapped with the need for current evidence), combined the patient and institutional context fields and no longer report the derived any-context field, leaving 20 fields in six blocks (Fig. 1b and Extended Data Table 1). Agreement for every annotated field is given in Extended Data Table 2. The guidelines require each field to be judged independently from the query text alone, without assuming facts that are not stated; surface features are coded literally; and a query with no recoverable request is coded as `Other' task and intent, low answerability and clarification. The same guideline text served as the clinicians' reference and as the model's instructions, and the two were verified to be byte-identical before every model run. The full text, including every definition, boundary rule and tie-break that the annotator applies to choose a category, is reproduced in Supplementary Note 1, and the request format and output schema are given in Supplementary Note 2.

\subsection{\textbf{Assessment of the task and intent categories}}

Because the task and intent categories were synthesized from several existing schemes rather than adopted from a single one, we assessed them in three ways. First, we mapped each category to its counterparts in established frameworks (Supplementary Table 1). Six task categories (drug information, treatment and management, diagnosis, test and result interpretation, procedural guidance and foundational knowledge) and most intents correspond to question types in pre-LLM clinical-question taxonomies, in which drug choice, the cause of a symptom and test selection were the commonest generic questions\textsuperscript{44,49,50}. The remaining categories (documentation and workflow, patient education and communication, coding and administrative, and the documentation-drafting and coding intents) correspond to the text-generation, communication and administrative tasks that frameworks for LLM evaluation and studies of LLM use added\textsuperscript{16,22,26}. Second, clinicians applied the categories reliably: Krippendorff's \ensuremath{\alpha} was 0.64 (95\% CI 0.55--0.73) for task and 0.45 (0.36--0.54) for intent (Extended Data Table 2), compared with \ensuremath{\kappa} = 0.53 when 11 coders classified clinical questions into Ely's 64 generic question types\textsuperscript{44}. Third, the categories separated queries with very different potential harm and safest response (Figs. 3b,c and 4d), the property RCQ-Map needs to serve as a stratification frame for evaluation.

\subsection{\textbf{Correspondence of RCQ-Map to prior literature}}

The risk, context and safest-response blocks adapt established frameworks, and the task and intent blocks correspond to earlier classifications of clinical questions, extended where generative AI introduces work that earlier frameworks did not anticipate.

\emph{Task category.} Classification by the deliverable the clinician wants corresponds to the generic clinical-question taxonomies developed from point-of-care questions, in which drug choice and dosing, the causes of symptoms and findings, test selection and management predominate\textsuperscript{44,49,51,52}. Keeping drug information and pharmacotherapy separate from treatment and management is consistent with these literatures, which identify drug questions as the single largest and most distinct class, and diagnosis, test and result interpretation and procedural guidance each map to a different knowledge source and error mode\textsuperscript{44,50,53}. Pre-LLM taxonomies contained no category for generating text; documentation and workflow, patient education and communication and coding and administrative tasks, which correspond to the note-generation, patient-communication and administration categories of the clinician-validated MedHELM taxonomy\textsuperscript{16} and to the writing tasks that dominate institutional and general-population LLM use\textsuperscript{22,26,27}. Because nurses and APPs raise different questions from physicians\textsuperscript{54,55}, categories were defined by task rather than by profession. An `Other' category captured non-clinical requests such as research writing and personal queries.

\emph{Question intent.} Consistent with the distinction in information-retrieval research between what a query is about and what the user is trying to achieve\textsuperscript{56}, intent records the motive behind the task: to verify a belief or planned action, look up a fact, reach a clinical decision, obtain a dose, draft text, understand a definition or mechanism, compare options, follow a procedure, obtain a code or interpret a result. These categories mirror the generic question stems of Ely et al. (for example, `What is the dose of drug x?' and `How should I manage condition x?')\textsuperscript{44} and the decision-oriented framing of evidence-based practice\textsuperscript{57}. Verification was defined separately because a clinician seeking confirmation has already formed a plan, and the decision to pursue a question is driven by its urgency and link to action\textsuperscript{43}. For comparisons, we grouped the 12 intents into five: decide or verify (clinical decision, verification, dosing and result interpretation), look up or understand (fact, definition, mechanism and comparison), draft text or code (documentation drafting and coding), follow a procedure, and other. Clinicians agreed more on the groups than on single intents (\ensuremath{\alpha} 0.61 versus 0.45, and 0.58--0.63 under alternative groupings).

\emph{Clinical department and division.} Rather than organ systems, clinical domain was assigned to the clinical department that would ordinarily own management of the problem, using a precedence ladder (for example, psychiatric content before pediatric, operative content to the operating department, drug questions to the department owning the treated condition), with a second field resolving the Department of Medicine into its divisions. Anchoring domains to accountable clinical owners allows query patterns to be directed to the services responsible for content governance, safety review and local guidance.

\emph{AMA use case.} To relate observed demand to the use cases that professional bodies track, each query was mapped to the AMA's list of physician AI use cases, as used in its physician surveys\textsuperscript{1}, with `None of the above' added because the list is not exhaustive.

\emph{Query properties.} Patient specificity distinguishes questions about an identifiable patient from general questions, a distinction central to information-needs research, in which unmet needs were predominantly patient-specific\textsuperscript{49,53}.

\emph{Context required.} Three binary fields identify what a safe answer needs that the query does not contain: patient-level information (history, medications, laboratory results), which clinicians most often lacked in observational studies\textsuperscript{53} and which EHR-grounded benchmarks supply\textsuperscript{58}; institutional knowledge (protocols, order sets, formularies, antibiograms, payer rules), which is local by definition; and current published evidence, reflecting the finding that answers are often absent from or out of date in the resources clinicians consult\textsuperscript{59} and motivating retrieval augmentation\textsuperscript{29,30}. Because clinicians rarely agreed on institutional knowledge as a separate field (\ensuremath{\alpha} 0.25), we report patient-level and institutional information together, as patient or local information that the query does not contain.

\emph{Potential harm.} Risk was defined as the potential harm if the query were answered incorrectly and the clinician acted on the answer, not as the probability that the model errs. The five grades (minimal, low, moderate, high, critical) adapt the severity levels of the AHRQ Common Formats Harm Scale (\url{https://www.psoppc.org)}, which has been used to rate the harm potential of LLM answers\textsuperscript{45,60}, together with the harm categories of the NCC MERP Index for Categorizing Medication Errors (\url{https://www.nccmerp.org/types-medication-errors}) and the WHO Conceptual Framework for the International Classification for Patient Safety (\url{https://www.who.int/publications/i/item/WHO-IER-PSP-2010.2)}. Anchors for the critical grade follow the ISMP list of high-alert medications (\url{https://home.ecri.org/blogs/ismp-resources/high-alert-medications-in-acute-care-settings)}, pediatric weight-based dosing\textsuperscript{61}, medication safety in older adults\textsuperscript{62}, teratogenic exposure and time-critical emergencies. Purely non-clinical harms (financial, legal, privacy) were capped at moderate.

\emph{Answerability and safest response.} Answerability records whether the query can be answered well as posed (high), only with material assumptions (partial) or not at all (low). The safest response (the `route' field in the annotation guidelines) operationalizes the selective-prediction and learning-to-defer literatures for a clinical assistant\textsuperscript{41,42,63}: answer directly; answer grounded in retrieved authoritative sources\textsuperscript{29,30,64}; ask the clinician for missing details\textsuperscript{31}; escalate to a qualified human such as a specialist or pharmacist\textsuperscript{32}; or abstain\textsuperscript{65}. We refer to escalation and abstention together as deferral. Responses are ordered by restrictiveness, and annotators select the least restrictive response that remains safe, with clarification taking precedence over retrieval when patient information is missing.

\emph{Surface features.} Nine literal features (medication, dose, laboratory or test result, imaging, vulnerable population, acute or urgent situation, request for text generation, grammatical form and use of clinical abbreviations) were coded independently of all judgment fields so that their association with risk and the safest response could be tested. Medication, dose and vulnerable-population cues reflect established drivers of medication harm\textsuperscript{61,62}; query form (question, command or fragment) follows query-log research showing that information-system queries are frequently keyword fragments rather than questions\textsuperscript{66,67}; and abbreviations were coded because clinical shorthand is highly ambiguous\textsuperscript{68}.

\subsection{\textbf{LLM annotation of the corpus}}

The corpus was annotated with GPT-5.6 sol (OpenAI; model identifier gpt-5.6-sol; \url{https://deploymentsafety.openai.com/gpt-5-6}) through the OpenAI Responses API with reasoning effort set to `none'. The annotation guidelines were supplied verbatim as the model's instructions (the system prompt) and each query as a single user message (`Question: \textless query text\textgreater'), with no chart data, user identity or conversation history (Supplementary Note 2). Output was constrained by a strict JSON schema enumerating the allowed values of all 24 fields, and every response was validated against the schema and the three hard consistency rules; invalid outputs and transient errors were retried with exponential backoff. LLM annotation of text corpora has been shown to match or exceed crowd annotation\textsuperscript{69}; because agreement with clinicians cannot be assumed in a clinical setting\textsuperscript{17,70}, we validated it directly.

\subsection{\textbf{Clinician validation study}}

A validation cohort of 100 queries was drawn at random (seed 42; Mulberry32 generator with a Fisher--Yates shuffle) from a de-identified extract of queries from the same deployment. Seven clinicians annotated the queries on a purpose-built web platform that displayed the full annotation guidelines with inline field rules, enforced the hard consistency rules, derived the any-context field automatically and required a value for every field. Queries were assigned breadth first so that each received exactly three independent annotations; each clinician received 40 queries initially and could request additional batches of 10 (range 30--60 per clinician). Grading clinicians could not see other clinicians' or information about the query sender. All 300 annotations were completed between 18 and 27 August 2026.

The same 100 queries were annotated on 24 August 2026 with the corpus annotator configuration: GPT-5.6 sol with reasoning effort `none' and the identical prompt and schema.

\textbf{Five benchmarks examined in detail}

To examine in detail where benchmarks depart from real use, we applied RCQ-Map to all released items of five public benchmarks downloaded from their official releases on 30 September 2026 (Extended Data Fig. 8 and Supplementary Table 2): the MedQA test set of United States Medical Licensing Examination-style questions (four-option version; 1,273 items)\textsuperscript{34}; HealthBench Professional (525 conversations)\textsuperscript{18}; Real-POCQi (620 questions)\textsuperscript{17}; and the two openly released subsets of MAST\textsuperscript{35}, NOHARM (330 items: 30 specialist eConsult cases, each with ten perturbed variants)\textsuperscript{37} and a script concordance test (174 items)\textsuperscript{36}. We selected these benchmarks purposively rather than by systematic search, to cover the two main ways clinical benchmarks are built: examination-style items, including MedQA as the most widely used\textsuperscript{11}, and questions or cases written by clinicians. The other MAST components are held out or require registration and were not included. Each item was presented to the annotator as the benchmark presents it to a model, reduced to a single request to match the unit of analysis for RCQs: the question stem followed by the lettered options for MedQA; the first user message for HealthBench Professional (115 of whose conversations continued beyond it); the question text for Real-POCQi; the case prompt for NOHARM; and the kit's test prompt (instructions, scenario, hypothesis and new information) for the script concordance test. Annotation used the corpus annotator configuration, guidelines, schema, validation and retry rules without modification. We summarized each source by its distributions of task, intent, potential harm and safest response and measured its distance from real use as the total variation distance between its distribution and that of the RCQ corpus, that is, half the sum of absolute differences across categories, which is 0 for identical distributions and 1 for distributions with no overlap. Confidence intervals for benchmarks are percentile bootstraps over items (2,000 resamples), or over base cases for NOHARM. Four of these benchmarks (all but Real-POCQi) are also in the Clinical AI Benchmark Atlas (below), where they are represented by samples of up to 200 items.

\subsection{\textbf{Clinical AI Benchmark Atlas}}

We assembled the benchmarks used to evaluate LLMs in health care from two sources. First, 13 public evaluation suites for medical LLMs, selected by the authors: MedHELM\textsuperscript{16}, MultiMedQA\textsuperscript{60}, MedS-Bench\textsuperscript{71}, ClinicBench\textsuperscript{72}, MIRAGE\textsuperscript{64}, Medmarks\textsuperscript{73}, BRIDGE\textsuperscript{74}, MEDIC\textsuperscript{75}, MMedBench\textsuperscript{76}, CLUE\textsuperscript{77}, MedBench\textsuperscript{78}, HealthBench\textsuperscript{15} and MAST\textsuperscript{35}. Each suite's component benchmarks were taken from its paper and code repository on 30 September 2026. Second, frontier model reports (technical reports and system or model cards of frontier models), selected from Epoch AI's Notable AI Models database (downloaded 4 October 2026)\textsuperscript{79}: every model with a language domain published between 1 January 2023 and 30 September 2026 by OpenAI, Anthropic, Google, Meta, Microsoft, xAI, Alibaba, DeepSeek, Mistral, Moonshot, Zhipu, MiniMax or NVIDIA. These 209 models link to 174 documents, of which 169 could be retrieved; the other five links led to a login page, a redirect, a single post or a withdrawn preprint. We read each document, together with any system card, model card or technical report from the same developer that it linked to, and recorded every health benchmark for which results were reported, with the passage reporting it. Twenty-three documents, covering 31 models, reported results on 22 health benchmarks.

Benchmark names from both sources were merged into a single list: alternative names of the same benchmark were combined, and subsets, translations and other variants were merged into their parent benchmark. A benchmark was included if (1) its evaluation items could be downloaded without credentialed access, an application, a data use agreement or payment (free registration was allowed); (2) its items, or an English version released by its authors, were in English; (3) its items could be answered from text alone (for mixed-modality benchmarks, the text-only items were used); (4) it was built for health care rather than drawn from a general-domain benchmark (MMLU, MMLU-Pro, GPQA, SuperGPQA, Humanity's Last Exam, TruthfulQA, C-Eval or CMMLU); and (5) its items were single requests, that is, questions, instructions, vignettes or conversations to which a model responds. Agent environments and token-level labeling tasks were recorded but not annotated. Criteria 1--3 and 5 were assessed on the downloaded items. Of 175 benchmarks named across the sources, 102 were excluded before download (42 not public, 40 not in English, 8 not answerable from text, 8 not built for health care and 4 not single requests) and 15 after download (8 not public, 4 not single requests, 2 versions of MedQA that were merged into it and 1 drawn from a general-domain benchmark), leaving 58 (Supplementary Data 1). For each benchmark we recorded the source, version (commit or dataset revision), file checksums, license and number of items.

From each benchmark, up to 200 items were sampled uniformly at random without replacement (seed 20261004) from its test or evaluation split, after removing items not in English, items that refer to an image that is not released, and duplicates; smaller benchmarks were used in full. Language was identified with lingua (version 2.1.1), and an item was treated as non-English when its top language was not English with a confidence of at least 0.80. Each item was rendered as a model under evaluation receives it: the question with its answer options; the vignette with its question; the instruction with its input text; the benchmark's or suite's published prompt template where one defines the task; or, for conversations, the first user turn, mirroring the first clinician turn annotated in the RCQ corpus. One benchmark whose published template leaves its slots unspecified could not be rendered as released and was excluded as not a single request.

RCQ-Map was applied to the 10,699 sampled items with the corpus annotator configuration (GPT-5.6 sol, reasoning effort none), guidelines, schema and retry rules. Of these, 10,659 (99.6\%) were labeled; the other 40, all from Med-HALT, failed the schema's consistency rules on every attempt. We summarized each benchmark separately and report medians and interquartile ranges across benchmarks for the 12 benchmarks used in frontier model reports, for the 29 benchmarks in the four suites whose papers state the aim of reflecting real-world clinical practice (MedHELM, BRIDGE, ClinicBench and HealthBench; MedBench, which states the same aim, contributed no eligible benchmark) and for all 58. Agreement with real use is reported as the share of the RCQ task mix that a benchmark covers, that is, one minus the total variation distance between its task distribution and that of the RCQ corpus; a distribution spread evenly across the ten task categories would cover 55\%. The 95\% confidence interval for the difference between the two groups' medians is from a percentile bootstrap that resampled benchmarks within each group (4,000 replicates).

\subsection{\textbf{Agreement statistics}}

Inter-rater reliability among clinicians was quantified for the 23 judged fields (the derived any-context field was excluded) with Krippendorff's \ensuremath{\alpha}, which accommodates the incomplete rater design, using the nominal metric for categorical fields and the ordinal metric for harm, answerability and safest response\textsuperscript{80,81}. Because chance-corrected coefficients are depressed for rare categories despite high raw agreement\textsuperscript{82}, we also report Gwet's AC1\textsuperscript{83}, Fleiss' \ensuremath{\kappa}\textsuperscript{84}, mean pairwise agreement and the share of unanimous queries. Following Krippendorff, \ensuremath{\alpha} \ensuremath{\geq} 0.800 indicates reliable and \ensuremath{\alpha} \ensuremath{\geq} 0.667 tentatively acceptable coding; other benchmarks are more lenient\textsuperscript{85}. Confidence intervals were obtained by bootstrapping queries (2,000 replicates). We also computed these statistics for the intent groups, the combined patient or local information field and two binary summaries, answerability below high and high or critical potential harm (Extended Data Table 2).

To compare the model with clinicians on equal terms, we used a held-out design. Because each query was annotated by a different subset of three clinicians, the held-out clinician rotated: for each query and each of its three clinicians, the reference was the label shared by the other two clinicians when they agreed; the held-out clinician's agreement with this reference was compared with the model's agreement with the same reference, and the difference was bootstrapped over queries (1,000 replicates). The model was never treated as an additional rater in reliability calculations. For ordinal fields we computed quadratic-weighted \ensuremath{\kappa}\textsuperscript{86} over all pairs of model and clinician ratings and all pairs of clinician ratings on the same queries. Consensus labels were defined as the majority label (at least two of three clinicians) or, for ordinal fields without a majority, the median grade; validation prevalences are reported with Wilson intervals\textsuperscript{87}.

To translate validation findings into corpus estimates, we estimated for each model-assigned grade the proportion of validation queries that the clinician majority placed in the target category (for example, high or critical risk) and applied these proportions to the corpus distribution of model-assigned grades. Confidence intervals were obtained by bootstrapping validation queries (2,000 replicates).

\subsection{\textbf{Validation results}}

Seven clinicians independently annotated a random sample of 100 RCQs with RCQ-Map, using the same annotation guidelines in a blinded, forced-choice interface, with three clinicians per RCQ (Fig. 1c). Clinicians agreed closely on literal surface features (Krippendorff's \ensuremath{\alpha} 0.55--0.90; Gwet's AC1 0.75--0.98) and on task category (\ensuremath{\alpha} 0.64, 95\% CI 0.55--0.73). They agreed less on the context an RCQ needed (\ensuremath{\alpha} 0.25--0.32), the safest response (0.40) and especially potential harm (ordinal \ensuremath{\alpha} 0.36, 95\% CI 0.23--0.48); all three clinicians chose the same harm grade for only 12 of the 100 RCQs and grades within one level of each other for 61 (Extended Data Fig. 6a and Extended Data Table 2). Individual clinicians rated between 0\% and 26\% of their RCQs as high or critical risk (Extended Data Fig. 6c). Grouping the 12 intents into five raised agreement on intent from \ensuremath{\alpha} 0.45 to 0.61, and agreement on answerability was 0.57 (Extended Data Table 2).

We then compared the corpus annotator (GPT-5.6 sol) with the clinicians against a common reference. Because each RCQ was annotated by a different set of three of the seven clinicians, we held out each of an RCQ's three clinicians in turn and took the label shared by the other two, when they agreed, as the reference; the model was compared with the same reference. Across the 23 judged fields, the model matched the reference in 84.6\% of comparisons and the held-out clinicians in 83.0\%. The model's agreement exceeded the clinician's for 14 of the 23 fields, significantly so for four (AMA use case, vulnerable population, text generation and query form), and was significantly lower for none (Extended Data Fig. 6b). For potential harm, the quadratic-weighted \ensuremath{\kappa} between the model and individual clinicians was 0.53, compared with 0.36 between pairs of clinicians, and 86\% of model ratings were within one grade of a clinician's rating, compared with 79\% of ratings by clinician pairs (Extended Data Fig. 6d). For the safest response, the weighted \ensuremath{\kappa} was 0.46 between the model and individual clinicians, 0.40 between pairs of clinicians and 0.53 between the model and clinician consensus (Extended Data Fig. 7). The model disagreed most where clinicians also disagreed: for task category, it matched the majority on 83\% of RCQs on which all three clinicians agreed and on 49\% of those on which they did not.

The model was more cautious about harm than the clinician majority. It rated 24\% of validation RCQs high or critical, compared with 6\% by the three-clinician majority (95\% CI 2.8\%--12.5\%); it flagged all six RCQs that the majority rated high or critical, but the majority agreed with only 25\% of its high- or critical-risk ratings (Extended Data Fig. 6c--e). Its rates of context requirements and of RCQs needing more than a direct answer matched the majority's. Calibrated to clinician consensus, the corpus prevalence of high- or critical-risk RCQs was 3.7\%, 95\% CI 1.2--5.9\% and the share needing more than a direct answer was 44.8\%, 95\% CI 35.7--54.4\%. We therefore treat 13.2\% as a cautious upper estimate of high-stakes use.

\subsection{\textbf{Statistical analysis of the corpus}}

Prevalences are reported as shares of annotated queries, with 95\% confidence intervals from a percentile cluster bootstrap that resampled clinicians (2,000 replicates), because conversations from the same clinician are correlated\textsuperscript{88}. Clinician-weighted estimates averaged per-clinician proportions. Sensitivity analyses excluded the 1\% of clinicians with the most conversations, excluded non-clinical (`Other') queries and stratified by professional group. The association between task mix and professional family is summarized by Cramér's V. Differences between professional families in the share of high- or critical-risk queries were tested by logistic regression with standard errors clustered by clinician. Concentration of use is summarized by the Lorenz curve and Gini coefficient.

Associations with high or critical potential harm (versus lower grades) were estimated by logistic regression with standard errors clustered by clinician\textsuperscript{89}. Model 1 included the surface features, patient specificity, query form and professional family; model 2 also included task category (reference, documentation and workflow). To assess whether risk can be recognized from the query text before answering, we fitted a logistic model with 11 text-observable features (eight binary surface features, two query-form contrasts and patient specificity), without task or professional family, in five-fold cross-validation grouped by clinician, and report the pooled out-of-fold AUC and the operating point that captures 90\% of high- or critical-risk queries.

For out-of-specialty analyses we mapped 31 organ- or discipline-defined specialties to the clinical department (and, for Department of Medicine subspecialties, the division) that owns their clinical territory (for example, cardiology to the Cardiology division of the Department of Medicine, and ophthalmology to Ophthalmology). Generalist specialties (internal medicine, hospital medicine, family medicine and emergency medicine) were excluded, as were queries with no clinical subject or with cross-cutting content (population health, biochemistry and molecular pharmacology, forensic medicine). A query was out of department if its assigned department was not among those mapped to the asker's specialty, and out of division if, in addition, its Medicine division did not match.

Analyses used Python 3.9 with pandas 2.3, statsmodels 0.13, scikit-learn 1.0 and matplotlib 3.5. Reporting follows the STROBE and TRIPOD-LLM guidelines where applicable\textsuperscript{90,91}.

\FloatBarrier
\section{\textbf{Data availability}}

Query text cannot be shared because it may contain protected health information and is subject to institutional agreement. The label-level dataset (24 annotation fields, professional family and suppressed specialty, without query text) is available from the corresponding author on reasonable request, subject to institutional approval and a data use agreement. Clinician and model labels for the validation set, without query text, are available from the corresponding author on reasonable request. The annotation guidelines are provided as Supplementary Note 1 and in the RCQ-Map repository (\url{https://github.com/nyuolab/RCQ-Map}; \url{https://doi.org/10.5281/zenodo.23223807)}. The Clinical AI Benchmark Atlas (CABA v1.0) lists each benchmark's source, version and license, together with the RCQ-Map labels for every sampled item. It is provided as Supplementary Data 1, archived at \url{https://doi.org/10.5281/zenodo.23223809} and released as a Hugging Face dataset (\url{https://huggingface.co/datasets/NYU-OLAB/CABA)}. Benchmark items are redistributed only where their licenses allow. All other items are listed by item ID and can be rebuilt from their original sources, under their own terms, with the scripts in the CABA repository. The aggregate RCQ distributions used for comparison are included with CABA.

\FloatBarrier
\section{\textbf{Code availability}}

The RCQ-Map annotation guidelines, output schema, prompts, annotation pipeline and clinician annotation platform are available at \url{https://github.com/nyuolab/RCQ-Map} (version 1.0, archived at \url{https://doi.org/10.5281/zenodo.23223807)}. Code to assemble CABA, sample and render benchmark items, apply RCQ-Map and compare benchmarks with real use is available at \url{https://github.com/nyuolab/CABA} (archived at \url{https://doi.org/10.5281/zenodo.23223809)}. The guidelines are released under a CC BY 4.0 license and the code under the Apache License 2.0.

\FloatBarrier
\section{\textbf{References}}

{\small\hangindent=1.5em\hangafter=1\noindent 1. American Medical Association. 2026 Physician Survey on Augmented Intelligence (AMA Center for Digital Health and AI, 2026); \url{https://www.ama-assn.org/system/files/physician-ai-sentiment-report.pdf}\par}\vspace{2pt}

{\small\hangindent=1.5em\hangafter=1\noindent 2. Shah, N. H., Entwistle, D. \& Pfeffer, M. A. Creation and adoption of large language models in medicine. JAMA 330, 866--869 (2023).\par}\vspace{2pt}

{\small\hangindent=1.5em\hangafter=1\noindent 3. Tierney, A. A. et al. Ambient artificial intelligence scribes to alleviate the burden of clinical documentation. NEJM Catal. Innov. Care Deliv. 5, CAT.23.0404 (2024).\par}\vspace{2pt}

{\small\hangindent=1.5em\hangafter=1\noindent 4. Garcia, P. et al. Artificial intelligence--generated draft replies to patient inbox messages. JAMA Netw. Open 7, e243201 (2024).\par}\vspace{2pt}

{\small\hangindent=1.5em\hangafter=1\noindent 5. Bedi, S. et al. Testing and evaluation of health care applications of large language models: a systematic review. JAMA 333, 319--328 (2025).\par}\vspace{2pt}

{\small\hangindent=1.5em\hangafter=1\noindent 6. Chen, S. F. et al. LLM-assisted systematic review of large language models in clinical medicine. Nat. Med. 32, 1152--1159 (2026).\par}\vspace{2pt}

{\small\hangindent=1.5em\hangafter=1\noindent 7. Raji, I. D., Daneshjou, R. \& Alsentzer, E. It's time to bench the medical exam benchmark. NEJM AI 2, AIe2401235 (2025).\par}\vspace{2pt}

{\small\hangindent=1.5em\hangafter=1\noindent 8. Nori, H., King, N., McKinney, S. M., Carignan, D. \& Horvitz, E. Capabilities of GPT-4 on medical challenge problems. Preprint at \url{https://arxiv.org/abs/2303.13375} (2023).\par}\vspace{2pt}

{\small\hangindent=1.5em\hangafter=1\noindent 9. Kung, T. H. et al. Performance of ChatGPT on USMLE: potential for AI-assisted medical education using large language models. PLOS Digit. Health 2, e0000198 (2023).\par}\vspace{2pt}

{\small\hangindent=1.5em\hangafter=1\noindent 10. Singhal, K. et al. Toward expert-level medical question answering with large language models. Nat. Med. 31, 943--950 (2025).\par}\vspace{2pt}

{\small\hangindent=1.5em\hangafter=1\noindent 11. Alaa, A. et al. Position: medical large language model benchmarks should prioritize construct validity. In Proc. 42nd International Conference on Machine Learning, PMLR 267, 80991--81004 (2025).\par}\vspace{2pt}

{\small\hangindent=1.5em\hangafter=1\noindent 12. Hager, P. et al. Evaluation and mitigation of the limitations of large language models in clinical decision-making. Nat. Med. 30, 2613--2622 (2024).\par}\vspace{2pt}

{\small\hangindent=1.5em\hangafter=1\noindent 13. Tu, T. et al. Towards conversational diagnostic artificial intelligence. Nature 642, 442--450 (2025).\par}\vspace{2pt}

{\small\hangindent=1.5em\hangafter=1\noindent 14. Johri, S. et al. An evaluation framework for clinical use of large language models in patient interaction tasks. Nat. Med. 31, 77--86 (2025).\par}\vspace{2pt}

{\small\hangindent=1.5em\hangafter=1\noindent 15. Arora, R. K. et al. HealthBench: evaluating large language models towards improved human health. Preprint at \url{https://arxiv.org/abs/2505.08775} (2025).\par}\vspace{2pt}

{\small\hangindent=1.5em\hangafter=1\noindent 16. Bedi, S. et al. Holistic evaluation of large language models for medical tasks with MedHELM. Nat. Med. 32, 943--951 (2026).\par}\vspace{2pt}

{\small\hangindent=1.5em\hangafter=1\noindent 17. Feng, J. et al. Expert evaluation of clinical AI tools on real point-of-care clinical queries. Preprint at \url{https://arxiv.org/abs/2606.28960} (2026).\par}\vspace{2pt}

{\small\hangindent=1.5em\hangafter=1\noindent 18. Hicks, R. S. et al. HealthBench Professional: evaluating large language models on real clinician chats. Preprint at \url{https://arxiv.org/abs/2604.27470} (2026).\par}\vspace{2pt}

{\small\hangindent=1.5em\hangafter=1\noindent 19. Ma, Y. et al. An evaluation framework for large language models in clinical nursing: a scoping review and expert consultation. J. Nurs. Manag. 2026, 9524014 (2026).\par}\vspace{2pt}

{\small\hangindent=1.5em\hangafter=1\noindent 20. Yu, S.-Y., Lin, H.-P., Kung, Y. M. \& Chen, S.-C. Artificial intelligence-enhanced clinical reasoning in nurse practitioners: a systematic review. Nurse Educ. Pract. 91, 104724 (2026).\par}\vspace{2pt}

{\small\hangindent=1.5em\hangafter=1\noindent 21. Bean, A. M. et al. Measuring what matters: construct validity in large language model benchmarks. In Advances in Neural Information Processing Systems 38, 19868--19949 (2025).\par}\vspace{2pt}

{\small\hangindent=1.5em\hangafter=1\noindent 22. Unell, A., Kashyap, M., Pfeffer, M. \& Shah, N. Real-world usage patterns of large language models in healthcare. Preprint at medRxiv \url{https://doi.org/10.1101/2025.05.02.25326781} (2025).\par}\vspace{2pt}

{\small\hangindent=1.5em\hangafter=1\noindent 23. Black, K. C. et al. Uses of generative AI by non-clinician staff at an academic medical center. npj Health Syst. 3, 13 (2026).\par}\vspace{2pt}

{\small\hangindent=1.5em\hangafter=1\noindent 24. Rader, B. et al. Utilization of a HIPAA-compliant large language model chatbot in an academic pediatric medical center. PLOS Digit. Health 5, e0001684 (2026).\par}\vspace{2pt}

{\small\hangindent=1.5em\hangafter=1\noindent 25. Shah, N. H. et al. Adoption and use of LLMs at an academic medical center. Preprint at \url{https://arxiv.org/abs/2602.00074} (2026).\par}\vspace{2pt}

{\small\hangindent=1.5em\hangafter=1\noindent 26. Tamkin, A. et al. Clio: privacy-preserving insights into real-world AI use. Preprint at \url{https://arxiv.org/abs/2412.13678} (2024).\par}\vspace{2pt}

{\small\hangindent=1.5em\hangafter=1\noindent 27. Handa, K. et al. Which economic tasks are performed with AI? Evidence from millions of Claude conversations. Preprint at \url{https://arxiv.org/abs/2503.04761} (2025).\par}\vspace{2pt}

{\small\hangindent=1.5em\hangafter=1\noindent 28. Chatterji, A. et al. How people use ChatGPT. NBER Working Paper 34255 (National Bureau of Economic Research, 2025).\par}\vspace{2pt}

{\small\hangindent=1.5em\hangafter=1\noindent 29. Lewis, P. et al. Retrieval-augmented generation for knowledge-intensive NLP tasks. Adv. Neural Inf. Process. Syst. 33, 9459--9474 (2020).\par}\vspace{2pt}

{\small\hangindent=1.5em\hangafter=1\noindent 30. Zakka, C. et al. Almanac --- retrieval-augmented language models for clinical medicine. NEJM AI 1, AIoa2300068 (2024).\par}\vspace{2pt}

{\small\hangindent=1.5em\hangafter=1\noindent 31. Li, S. S. et al. MediQ: question-asking LLMs and a benchmark for reliable interactive clinical reasoning. Adv. Neural Inf. Process. Syst. 37, 28858--28888 (2024).\par}\vspace{2pt}

{\small\hangindent=1.5em\hangafter=1\noindent 32. Dvijotham, K. et al. Enhancing the reliability and accuracy of AI-enabled diagnosis via complementarity-driven deferral to clinicians. Nat. Med. 29, 1814--1820 (2023).\par}\vspace{2pt}

{\small\hangindent=1.5em\hangafter=1\noindent 33. Finlayson, S. G. et al. The clinician and dataset shift in artificial intelligence. N. Engl. J. Med. 385, 283--286 (2021).\par}\vspace{2pt}

{\small\hangindent=1.5em\hangafter=1\noindent 34. Jin, D. et al. What disease does this patient have? A large-scale open domain question answering dataset from medical exams. Appl. Sci. 11, 6421 (2021).\par}\vspace{2pt}

{\small\hangindent=1.5em\hangafter=1\noindent 35. Goh, E. et al. Toward a test of medical AI superintelligence. Nat. Med. \url{https://doi.org/10.1038/s41591-026-04539-8} (2026).\par}\vspace{2pt}

{\small\hangindent=1.5em\hangafter=1\noindent 36. McCoy, L. G. et al. Assessment of large language models in clinical reasoning: a novel benchmarking study. NEJM AI 2, AIdbp2500120 (2025).\par}\vspace{2pt}

{\small\hangindent=1.5em\hangafter=1\noindent 37. Wu, D. et al. First, do NOHARM: a medical safety benchmark and randomized study of physician and AI teaming on clinical consultations. Preprint at \url{https://arxiv.org/abs/2512.01241} (2025).\par}\vspace{2pt}

{\small\hangindent=1.5em\hangafter=1\noindent 38. Sinsky, C. et al. Allocation of physician time in ambulatory practice: a time and motion study in 4 specialties. Ann. Intern. Med. 165, 753--760 (2016).\par}\vspace{2pt}

{\small\hangindent=1.5em\hangafter=1\noindent 39. Arndt, B. G. et al. Tethered to the EHR: primary care physician workload assessment using EHR event log data and time-motion observations. Ann. Fam. Med. 15, 419--426 (2017).\par}\vspace{2pt}

{\small\hangindent=1.5em\hangafter=1\noindent 40. Hendrich, A., Chow, M. P., Skierczynski, B. A. \& Lu, Z. A 36-hospital time and motion study: how do medical-surgical nurses spend their time? Perm. J. 12, 25--34 (2008).\par}\vspace{2pt}

{\small\hangindent=1.5em\hangafter=1\noindent 41. Geifman, Y. \& El-Yaniv, R. Selective classification for deep neural networks. Adv. Neural Inf. Process. Syst. 30 (2017).\par}\vspace{2pt}

{\small\hangindent=1.5em\hangafter=1\noindent 42. Kamath, A., Jia, R. \& Liang, P. Selective question answering under domain shift. In Proc. 58th Annual Meeting of the Association for Computational Linguistics 5684--5696 (Association for Computational Linguistics, 2020).\par}\vspace{2pt}

{\small\hangindent=1.5em\hangafter=1\noindent 43. Gorman, P. N. \& Helfand, M. Information seeking in primary care: how physicians choose which clinical questions to pursue and which to leave unanswered. Med. Decis. Making 15, 113--119 (1995).\par}\vspace{2pt}

{\small\hangindent=1.5em\hangafter=1\noindent 44. Ely, J. W. et al. A taxonomy of generic clinical questions: classification study. BMJ 321, 429--432 (2000).\par}\vspace{2pt}

{\small\hangindent=1.5em\hangafter=1\noindent 45. Williams, T., Szekendi, M., Pavkovic, S., Clevenger, W. \& Cerese, J. The reliability of AHRQ Common Format Harm Scales in rating patient safety events. J. Patient Saf. 11, 52--59 (2015).\par}\vspace{2pt}

{\small\hangindent=1.5em\hangafter=1\noindent 46. Sweeney, L. k-anonymity: a model for protecting privacy. Int. J. Uncertain. Fuzziness Knowl. Based Syst. 10, 557--570 (2002).\par}\vspace{2pt}

{\small\hangindent=1.5em\hangafter=1\noindent 47. El Emam, K. \& Dankar, F. K. Protecting privacy using k-anonymity. J. Am. Med. Inform. Assoc. 15, 627--637 (2008).\par}\vspace{2pt}

{\small\hangindent=1.5em\hangafter=1\noindent 48. Vishwanath, K. et al. General-purpose large language models outperform specialized clinical AI tools on medical benchmarks. Nat. Med. \url{https://doi.org/10.1038/s41591-026-04431-5} (2026).\par}\vspace{2pt}

{\small\hangindent=1.5em\hangafter=1\noindent 49. Ely, J. W. et al. Analysis of questions asked by family doctors regarding patient care. BMJ 319, 358--361 (1999).\par}\vspace{2pt}

{\small\hangindent=1.5em\hangafter=1\noindent 50. Allen, M. et al. The classification of clinicians' information needs while using a clinical information system. AMIA Annu. Symp. Proc. 2003, 26--30 (2003).\par}\vspace{2pt}

{\small\hangindent=1.5em\hangafter=1\noindent 51. Del Fiol, G., Workman, T. E. \& Gorman, P. N. Clinical questions raised by clinicians at the point of care: a systematic review. JAMA Intern. Med. 174, 710--718 (2014).\par}\vspace{2pt}

{\small\hangindent=1.5em\hangafter=1\noindent 52. Smith, R. What clinical information do doctors need? BMJ 313, 1062--1068 (1996).\par}\vspace{2pt}

{\small\hangindent=1.5em\hangafter=1\noindent 53. Currie, L. M. et al. Clinical information needs in context: an observational study of clinicians while using a clinical information system. AMIA Annu. Symp. Proc. 2003, 190--194 (2003).\par}\vspace{2pt}

{\small\hangindent=1.5em\hangafter=1\noindent 54. Cogdill, K. W. Information needs and information seeking in primary care: a study of nurse practitioners. J. Med. Libr. Assoc. 91, 203--215 (2003).\par}\vspace{2pt}

{\small\hangindent=1.5em\hangafter=1\noindent 55. McKnight, M. The information seeking of on-duty critical care nurses: evidence from participant observation and in-context interviews. J. Med. Libr. Assoc. 94, 145--151 (2006).\par}\vspace{2pt}

{\small\hangindent=1.5em\hangafter=1\noindent 56. Broder, A. A taxonomy of web search. SIGIR Forum 36, 3--10 (2002).\par}\vspace{2pt}

{\small\hangindent=1.5em\hangafter=1\noindent 57. Richardson, W. S., Wilson, M. C., Nishikawa, J. \& Hayward, R. S. The well-built clinical question: a key to evidence-based decisions. ACP J. Club 123, A12--A13 (1995).\par}\vspace{2pt}

{\small\hangindent=1.5em\hangafter=1\noindent 58. Fleming, S. L. et al. MedAlign: a clinician-generated dataset for instruction following with electronic medical records. Proc. AAAI Conf. Artif. Intell. 38, 22021--22030 (2024).\par}\vspace{2pt}

{\small\hangindent=1.5em\hangafter=1\noindent 59. Ely, J. W., Osheroff, J. A., Chambliss, M. L., Ebell, M. H. \& Rosenbaum, M. E. Answering physicians' clinical questions: obstacles and potential solutions. J. Am. Med. Inform. Assoc. 12, 217--224 (2005).\par}\vspace{2pt}

{\small\hangindent=1.5em\hangafter=1\noindent 60. Singhal, K. et al. Large language models encode clinical knowledge. Nature 620, 172--180 (2023).\par}\vspace{2pt}

{\small\hangindent=1.5em\hangafter=1\noindent 61. Kaushal, R. et al. Medication errors and adverse drug events in pediatric inpatients. JAMA 285, 2114--2120 (2001).\par}\vspace{2pt}

{\small\hangindent=1.5em\hangafter=1\noindent 62. 2023 American Geriatrics Society Beers Criteria\textregistered{} Update Expert Panel. American Geriatrics Society 2023 updated AGS Beers Criteria\textregistered{} for potentially inappropriate medication use in older adults. J. Am. Geriatr. Soc. 71, 2052--2081 (2023).\par}\vspace{2pt}

{\small\hangindent=1.5em\hangafter=1\noindent 63. Madras, D., Pitassi, T. \& Zemel, R. Predict responsibly: improving fairness and accuracy by learning to defer. Adv. Neural Inf. Process. Syst. 31 (2018).\par}\vspace{2pt}

{\small\hangindent=1.5em\hangafter=1\noindent 64. Xiong, G., Jin, Q., Lu, Z. \& Zhang, A. Benchmarking retrieval-augmented generation for medicine. In Findings of the Association for Computational Linguistics: ACL 2024 6233--6251 (Association for Computational Linguistics, 2024).\par}\vspace{2pt}

{\small\hangindent=1.5em\hangafter=1\noindent 65. Feng, S. et al. Don't hallucinate, abstain: identifying LLM knowledge gaps via multi-LLM collaboration. In Proc. 62nd Annual Meeting of the Association for Computational Linguistics (Volume 1: Long Papers) 14664--14690 (Association for Computational Linguistics, 2024).\par}\vspace{2pt}

{\small\hangindent=1.5em\hangafter=1\noindent 66. Jansen, B. J. \& Spink, A. How are we searching the World Wide Web? A comparison of nine search engine transaction logs. Inf. Process. Manag. 42, 248--263 (2006).\par}\vspace{2pt}

{\small\hangindent=1.5em\hangafter=1\noindent 67. Natarajan, K., Stein, D., Jain, S. \& Elhadad, N. An analysis of clinical queries in an electronic health record search utility. Int. J. Med. Inform. 79, 515--522 (2010).\par}\vspace{2pt}

{\small\hangindent=1.5em\hangafter=1\noindent 68. Moon, S., Pakhomov, S., Liu, N., Ryan, J. O. \& Melton, G. B. A sense inventory for clinical abbreviations and acronyms created using clinical notes and medical dictionary resources. J. Am. Med. Inform. Assoc. 21, 299--307 (2014).\par}\vspace{2pt}

{\small\hangindent=1.5em\hangafter=1\noindent 69. Gilardi, F., Alizadeh, M. \& Kubli, M. ChatGPT outperforms crowd workers for text-annotation tasks. Proc. Natl Acad. Sci. USA 120, e2305016120 (2023).\par}\vspace{2pt}

{\small\hangindent=1.5em\hangafter=1\noindent 70. Zheng, L. et al. Judging LLM-as-a-judge with MT-Bench and Chatbot Arena. Adv. Neural Inf. Process. Syst. 36, 46595--46623 (2023).\par}\vspace{2pt}

{\small\hangindent=1.5em\hangafter=1\noindent 71. Wu, C. et al. Towards evaluating and building versatile large language models for medicine. npj Digit. Med. 8, 58 (2025).\par}\vspace{2pt}

{\small\hangindent=1.5em\hangafter=1\noindent 72. Liu, F. et al. Large language models in the clinic: a comprehensive benchmark. Preprint at \url{https://arxiv.org/abs/2405.00716} (2024).\par}\vspace{2pt}

{\small\hangindent=1.5em\hangafter=1\noindent 73. Warner, B. et al. Medmarks: a comprehensive open-source LLM benchmark suite for medical tasks. Preprint at \url{https://arxiv.org/abs/2605.01417} (2026).\par}\vspace{2pt}

{\small\hangindent=1.5em\hangafter=1\noindent 74. Wu, J. et al. BRIDGE: benchmarking large language models for understanding real-world clinical practice text. Nat. Biomed. Eng. \url{https://doi.org/10.1038/s41551-026-01719-2} (2026).\par}\vspace{2pt}

{\small\hangindent=1.5em\hangafter=1\noindent 75. Kanithi, P. K. et al. MEDIC: comprehensive evaluation of leading indicators for LLM safety and utility in clinical applications. Preprint at \url{https://arxiv.org/abs/2409.07314} (2024).\par}\vspace{2pt}

{\small\hangindent=1.5em\hangafter=1\noindent 76. Qiu, P. et al. Towards building multilingual language model for medicine. Nat. Commun. 15, 8384 (2024).\par}\vspace{2pt}

{\small\hangindent=1.5em\hangafter=1\noindent 77. Goodwin, T. R. \& Demner-Fushman, D. Clinical Language Understanding Evaluation (CLUE). Preprint at \url{https://arxiv.org/abs/2209.14377} (2022).\par}\vspace{2pt}

{\small\hangindent=1.5em\hangafter=1\noindent 78. Cai, Y. et al. MedBench: a large-scale Chinese benchmark for evaluating medical large language models. Proc. AAAI Conf. Artif. Intell. 38, 17709--17717 (2024).\par}\vspace{2pt}

{\small\hangindent=1.5em\hangafter=1\noindent 79. Epoch AI. Data on notable AI models. \url{https://epoch.ai/data/notable-ai-models} (accessed 4 October 2026).\par}\vspace{2pt}

{\small\hangindent=1.5em\hangafter=1\noindent 80. Hayes, A. F. \& Krippendorff, K. Answering the call for a standard reliability measure for coding data. Commun. Methods Meas. 1, 77--89 (2007).\par}\vspace{2pt}

{\small\hangindent=1.5em\hangafter=1\noindent 81. Krippendorff, K. Content Analysis: An Introduction to Its Methodology 4th edn (SAGE, 2018).\par}\vspace{2pt}

{\small\hangindent=1.5em\hangafter=1\noindent 82. Feinstein, A. R. \& Cicchetti, D. V. High agreement but low kappa: I. The problems of two paradoxes. J. Clin. Epidemiol. 43, 543--549 (1990).\par}\vspace{2pt}

{\small\hangindent=1.5em\hangafter=1\noindent 83. Gwet, K. L. Computing inter-rater reliability and its variance in the presence of high agreement. Br. J. Math. Stat. Psychol. 61, 29--48 (2008).\par}\vspace{2pt}

{\small\hangindent=1.5em\hangafter=1\noindent 84. Fleiss, J. L. Measuring nominal scale agreement among many raters. Psychol. Bull. 76, 378--382 (1971).\par}\vspace{2pt}

{\small\hangindent=1.5em\hangafter=1\noindent 85. Landis, J. R. \& Koch, G. G. The measurement of observer agreement for categorical data. Biometrics 33, 159--174 (1977).\par}\vspace{2pt}

{\small\hangindent=1.5em\hangafter=1\noindent 86. Cohen, J. Weighted kappa: nominal scale agreement with provision for scaled disagreement or partial credit. Psychol. Bull. 70, 213--220 (1968).\par}\vspace{2pt}

{\small\hangindent=1.5em\hangafter=1\noindent 87. Wilson, E. B. Probable inference, the law of succession, and statistical inference. J. Am. Stat. Assoc. 22, 209--212 (1927).\par}\vspace{2pt}

{\small\hangindent=1.5em\hangafter=1\noindent 88. Field, C. A. \& Welsh, A. H. Bootstrapping clustered data. J. R. Stat. Soc. B 69, 369--390 (2007).\par}\vspace{2pt}

{\small\hangindent=1.5em\hangafter=1\noindent 89. Cameron, A. C. \& Miller, D. L. A practitioner's guide to cluster-robust inference. J. Hum. Resour. 50, 317--372 (2015).\par}\vspace{2pt}

{\small\hangindent=1.5em\hangafter=1\noindent 90. von Elm, E. et al. The Strengthening the Reporting of Observational Studies in Epidemiology (STROBE) statement: guidelines for reporting observational studies. Lancet 370, 1453--1457 (2007).\par}\vspace{2pt}

{\small\hangindent=1.5em\hangafter=1\noindent 91. Gallifant, J. et al. The TRIPOD-LLM reporting guideline for studies using large language models. Nat. Med. 31, 60--69 (2025).\par}\vspace{2pt}

\FloatBarrier
\section{\textbf{Acknowledgements}}

We thank the health-system IT, privacy and data teams who processed the corpus, and Bryce McDonnell and Walter Wang for helping with data access, egress, and execution. We acknowledge Nader Mherabi and Dafna Bar-Sagi, Ph.D., for their support of medical AI research at NYU Langone. We thank M. Constantino and the NYULH High-Performance Computing (HPC) Team for computing resources essential to our work.

\FloatBarrier
\section{\textbf{Author contributions}}

TR and EKO supervised the study. KV and EKO conceptualized and established the study design. KV designed and administered the clinician validation of RCQ-Map. JVL, DBH, JJY, WRS, HAK, CO, and AV annotated the validation set. TR, HL, KV, and YA prepared and acquired the data. KV performed the statistical analysis. KV and TR contributed to study evaluation and data review. KV wrote the initial draft and developed the figures of the manuscript. All authors reviewed and approved the final manuscript.

\FloatBarrier
\section{\textbf{Competing interests}}

EKO reports equity in MarchAI and Artisight, spousal employment by Eikon Therapeutics, and consulting for Sofinnova Partners and Google. KV reports consulting for ChartR Health. The remaining authors declare no competing interests.

\FloatBarrier
\section{\textbf{Funding}}

EKO is supported by the National Cancer Institute's Early-Stage Surgeon Scientist Program (3P30CA016087-41S1) and the W.M. Keck Foundation. This work was supported by the Institute for Information \& Communications Technology Planning and Evaluation (IITP) grant funded by the Ministry of Science and ICT (MSIT) of the Republic of Korea government (No. RS-2019-II190075 Artificial Intelligence Graduate School Program (KAIST); No. RS-2024-00509279, Global AI Frontier Lab).

\clearpage

\FloatBarrier
\section{\textbf{Extended Data}}

\begin{figure}[!htbp]
\centering
\includegraphics[width=\textwidth,height=0.85\textheight,keepaspectratio]{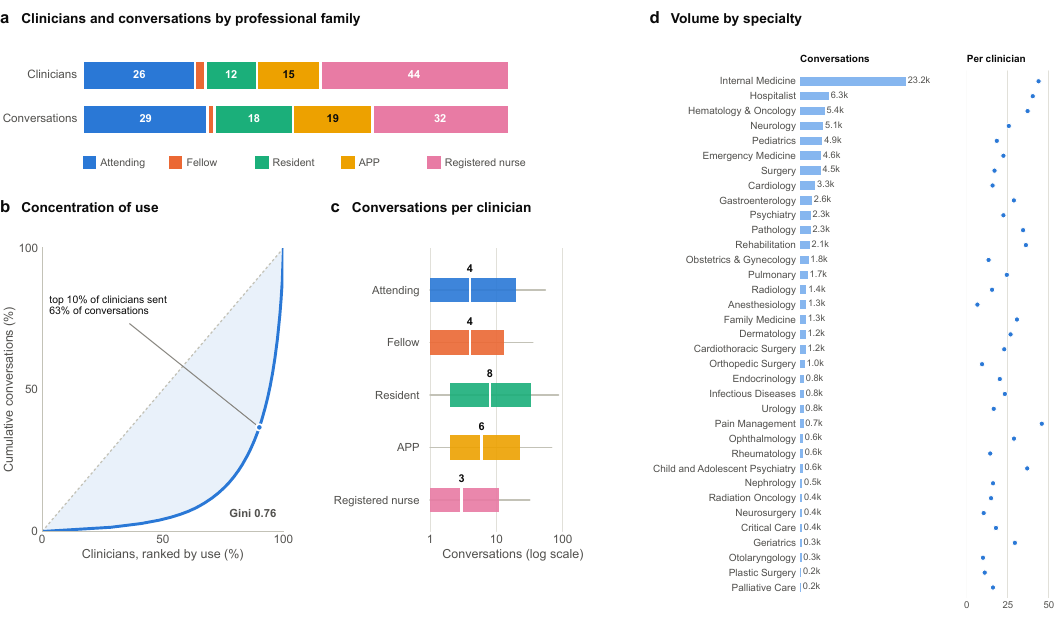}
\par\vspace{6pt}\begingroup\small\raggedright
\textbf{Extended Data Fig. 1 \textbar{} Usage patterns.} \textbf{a}, Share of clinicians and of conversations by professional family. \textbf{b}, Lorenz curve of conversations across clinicians ranked by use; the shaded area is proportional to the Gini coefficient. \textbf{c,} Conversations per clinician by professional family (log scale); bars, interquartile range; lines, 10th--90th percentile; numbers, median. \textbf{d}, Conversations and conversations per clinician for the 35 released specialties (physicians and APPs), all of which had at least 200 conversations.\par\endgroup
\end{figure}

\begin{figure}[!htbp]
\centering
\includegraphics[width=\textwidth,height=0.85\textheight,keepaspectratio]{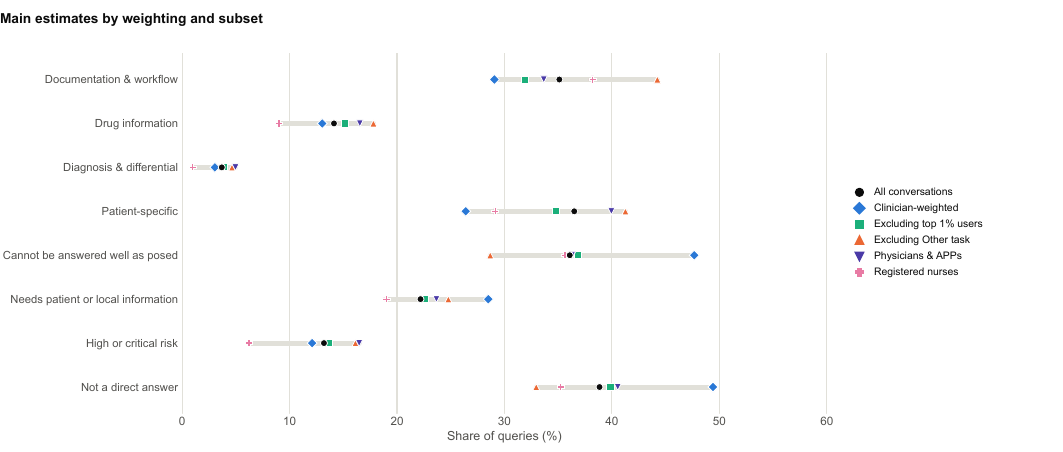}
\par\vspace{6pt}\begingroup\small\raggedright
\textbf{Extended Data Fig. 2 \textbar{} Robustness of headline estimates.} Estimates for all conversations, weighting each clinician equally, excluding the 1\% of clinicians with the most conversations, excluding non-clinical (`Other') queries, and for physicians and APPs and for RNs separately. Gray bands span the range across variants.\par\endgroup
\end{figure}

\begin{figure}[!htbp]
\centering
\includegraphics[width=\textwidth,height=0.85\textheight,keepaspectratio]{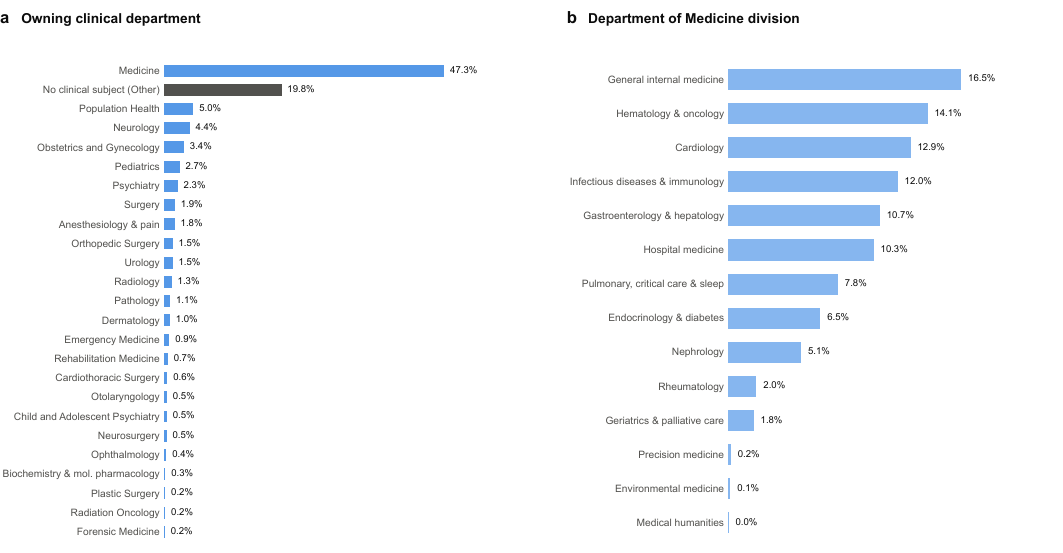}
\par\vspace{6pt}\begingroup\small\raggedright
\textbf{Extended Data Fig. 3 \textbar{} Clinical department and Department of Medicine division.} \textbf{a}, Share of all queries by the department that would ordinarily own management of the problem, assigned by the precedence ladder in the annotation guidelines; `No clinical subject' covers queries without clinical content. \textbf{b}, Division within the Department of Medicine, as a share of Medicine queries.\par\endgroup
\end{figure}

\begin{figure}[!htbp]
\centering
\includegraphics[width=\textwidth,height=0.85\textheight,keepaspectratio]{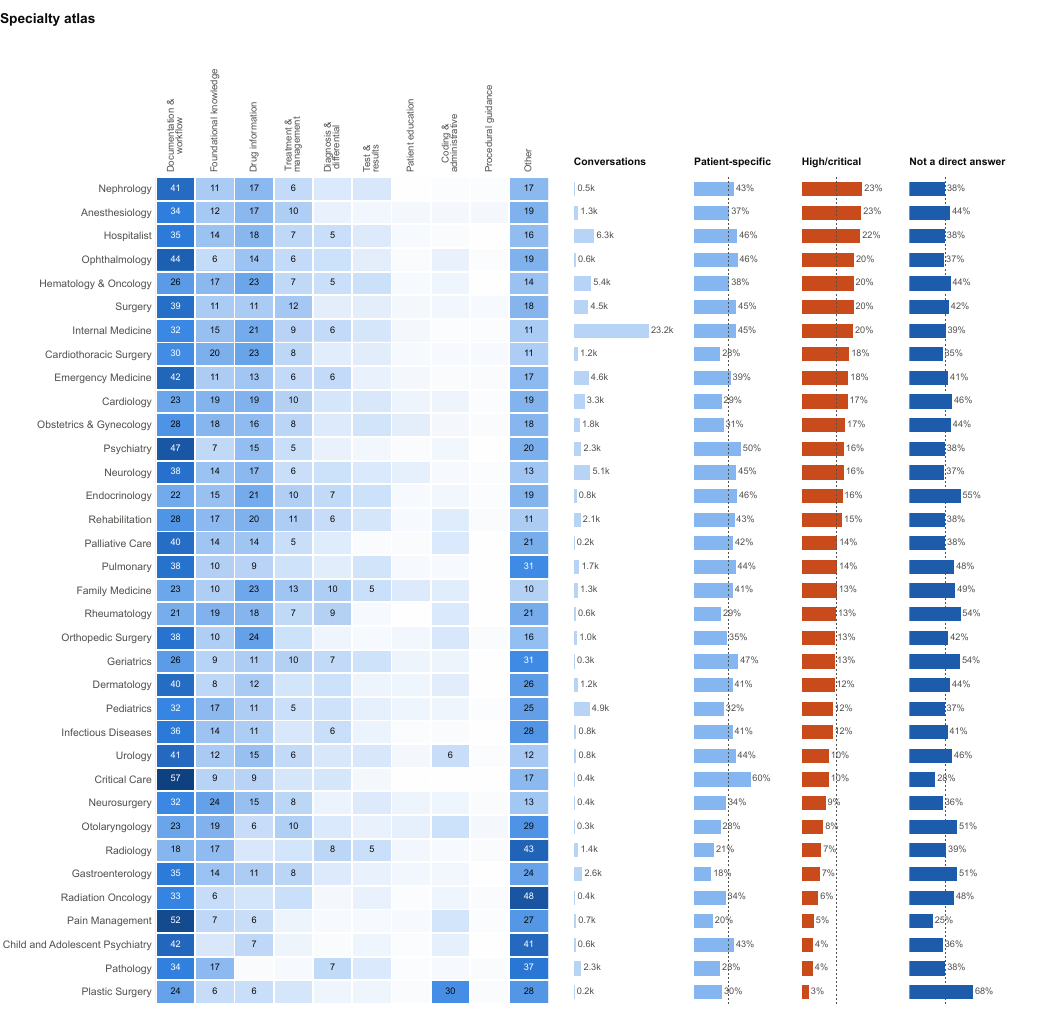}
\par\vspace{6pt}\begingroup\small\raggedright
\textbf{Extended Data Fig. 4 \textbar{} Specialty atlas.} Task mix (\% of each specialty's queries; cells \textless5\% unlabeled), conversation volume, share of patient-specific queries, share of high- or critical-risk queries and share of queries whose safest response is not a direct answer, for the 35 released specialties, sorted by high- or critical-risk share. Dashed lines, all-query values.\par\endgroup
\end{figure}

\begin{figure}[!htbp]
\centering
\includegraphics[width=\textwidth,height=0.85\textheight,keepaspectratio]{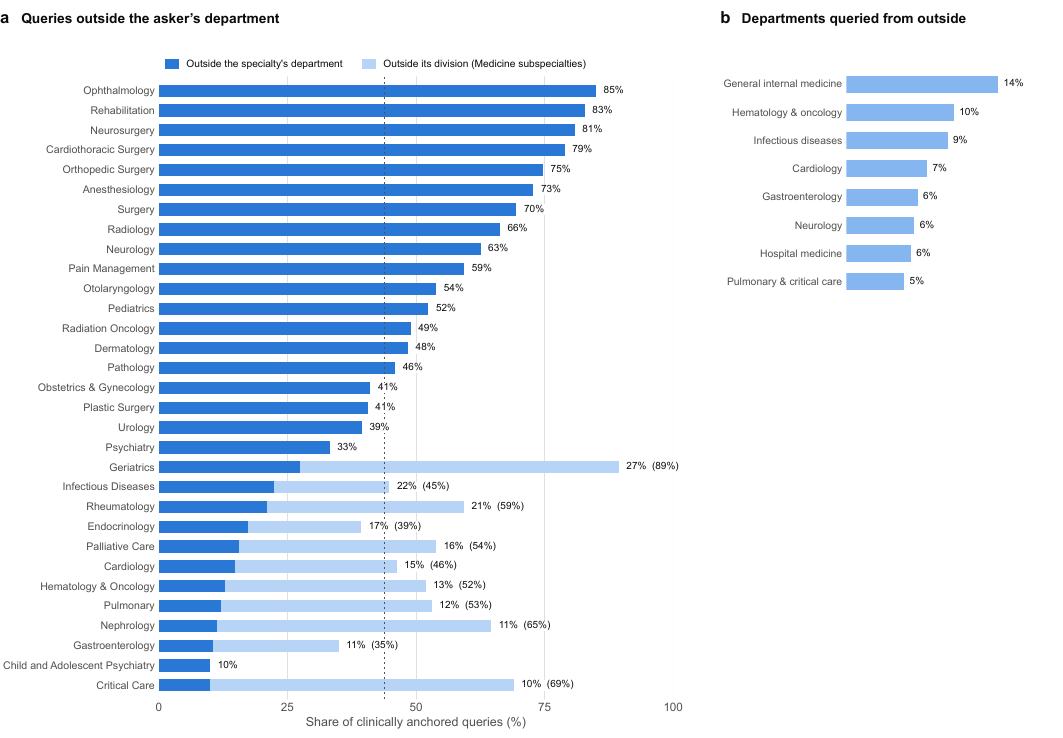}
\par\vspace{6pt}\begingroup\small\raggedright
\textbf{Extended Data Fig. 5 \textbar{} Queries outside the asker's specialty.} \textbf{a}, Share of clinically anchored queries from each organ- or discipline-defined specialty that concerned a condition owned by another department (dark) or, for Department of Medicine subspecialties, another division (light; in parentheses). Dashed line, all specialties. Generalist specialties and queries without clinical or with cross-cutting content are excluded (Methods). \textbf{b}, The departments and divisions most often queried from outside, as shares of out-of-division queries.\par\endgroup
\end{figure}

\begin{figure}[!htbp]
\centering
\includegraphics[width=\textwidth,height=0.71\textheight,keepaspectratio]{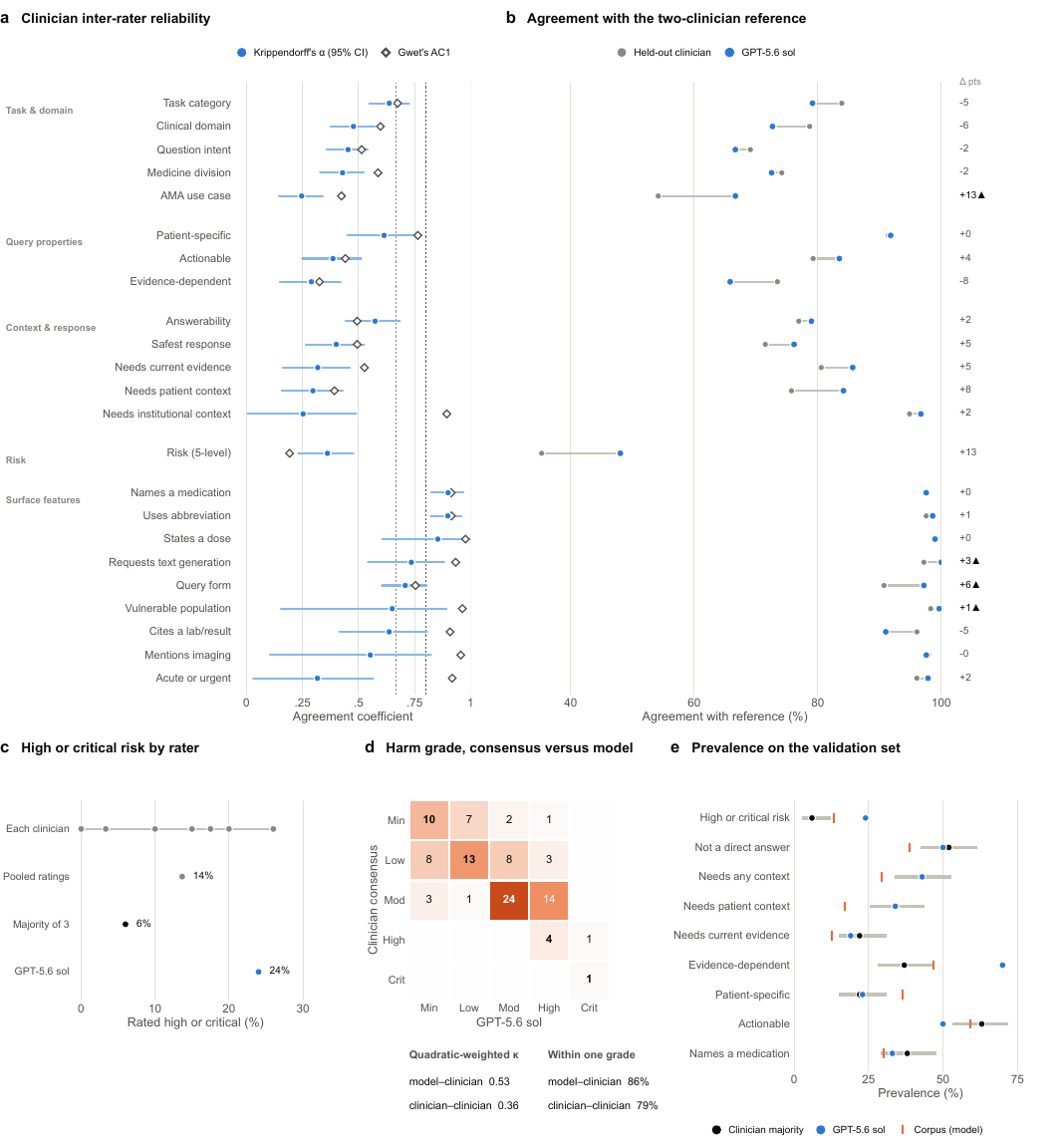}
\par\vspace{6pt}\begingroup\small\raggedright
\textbf{Extended Data Fig. 6 \textbar{} Clinician validation of RCQ-Map annotation.} One hundred randomly sampled queries, each annotated independently by three of seven clinicians (300 annotations) and by the corpus annotator (GPT-5.6 sol) with the same annotation guidelines. \textbf{a}, Inter-rater reliability for the 23 judged fields: Krippendorff's \ensuremath{\alpha} (ordinal metric for harm, answerability and safest response) with 95\% bootstrap CIs, and Gwet's AC1; dashed lines, \ensuremath{\alpha} = 0.667 and 0.800. \textbf{b}, Agreement with a two-clinician reference (the label shared by two clinicians when they agree) for the held-out third clinician (gray) and for GPT-5.6 sol (blue); right, difference in percentage points, with \ensuremath{\blacktriangle} or \ensuremath{\blacktriangledown} where the paired bootstrap 95\% CI excludes zero. \textbf{c}, Share of queries rated high or critical by each clinician (gray points; each clinician rated 30--60 queries), across all clinician ratings, by the three-clinician majority and by GPT-5.6 sol. \textbf{d}, Harm grade by clinician consensus (majority, or median grade when all three differed) versus GPT-5.6 sol, with quadratic-weighted \ensuremath{\kappa} and within-one-grade agreement for pairs of model and clinician ratings and for pairs of clinician ratings. \textbf{e}, Prevalence of selected labels on the validation set by clinician majority (95\% Wilson CI) and by GPT-5.6 sol, with the model-annotated corpus prevalence for reference.\par\endgroup
\end{figure}

\begin{figure}[!htbp]
\centering
\includegraphics[width=\textwidth,height=0.85\textheight,keepaspectratio]{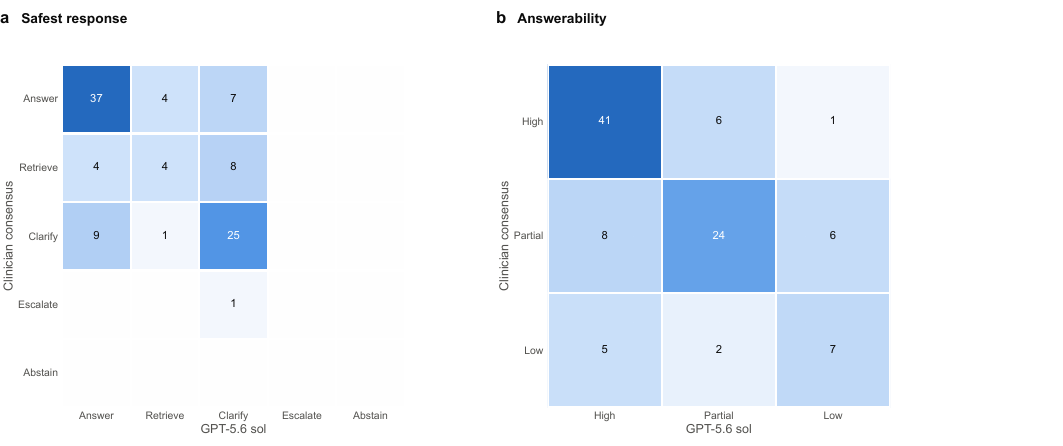}
\par\vspace{6pt}\begingroup\small\raggedright
\textbf{Extended Data Fig. 7 \textbar{} Safest response and answerability: clinician consensus versus the model.} Confusion matrices between clinician consensus (majority, or median when all three differed) and GPT-5.6 sol for the safest response (\textbf{a}) and answerability (\textbf{b}) on the 100 validation queries. Quadratic-weighted \ensuremath{\kappa} between consensus and the model was 0.53 for the safest response and 0.55 for answerability (exact agreement 66\% and 72\%).\par\endgroup
\end{figure}

\begin{figure}[!htbp]
\centering
\includegraphics[width=\textwidth,height=0.67\textheight,keepaspectratio]{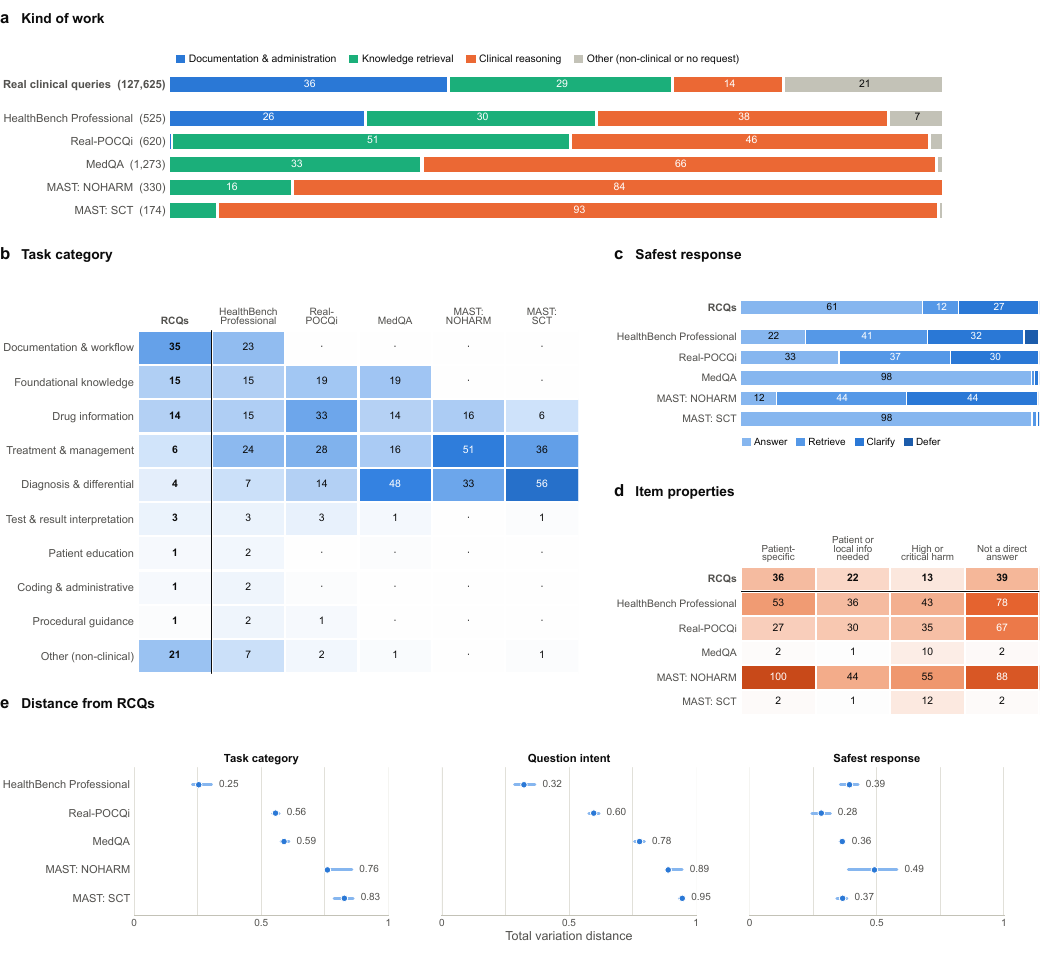}
\par\vspace{6pt}\begingroup\small\raggedright
\textbf{Extended Data Fig. 8 \textbar{} Five public benchmarks compared with real clinical use.} Real clinical queries (RCQs; n = 127,625) and five public benchmarks, all annotated with RCQ-Map by the same annotator (GPT-5.6 sol, reasoning effort none) with the same guidelines and schema. Real-POCQi, questions that physicians submitted to a clinical AI tool at the point of care; HealthBench Professional, first user turns of conversations written by physicians; MAST: NOHARM, specialist eConsult cases and their perturbed variants; MAST: SCT, script concordance test items; MedQA, examination questions in the style of the United States Medical Licensing Examination (Methods). Benchmarks are ordered from closest to farthest from RCQs in task distribution (e). \textbf{a}, Kind of work requested, with task categories grouped as in Fig. 2a; numbers of items in parentheses. \textbf{b}, Task category (\% of each source's items; dots, none). \textbf{c}, Safest response: answer directly, retrieve sources and then answer, ask the clinician to clarify, or defer (escalate to a person or abstain). \textbf{d}, Shares of items that concern a specific patient, need context that the item does not contain, carry high or critical potential harm, or are safest met by something other than a direct answer. \textbf{e}, Total variation distance between each benchmark's distribution and the RCQ distribution for task, intent and safest response (0, identical; 1, no overlap), with 95\% bootstrap CIs over items (over base cases for NOHARM). Full distributions are given in Supplementary Table 2.\par\endgroup
\end{figure}

\FloatBarrier\Needspace*{18\baselineskip}\noindent{\small \textbf{Extended Data Table 1 \textbar{} RCQ-Map fields, allowed values and grounding in prior literature.}\par}\vspace{4pt}
\begingroup\small\setlength{\tabcolsep}{3pt}\renewcommand{\arraystretch}{1.12}
\begin{longtable}{@{}>{\raggedright\arraybackslash}p{0.1335\dimexpr\textwidth-8\tabcolsep\relax}>{\raggedright\arraybackslash}p{0.2297\dimexpr\textwidth-8\tabcolsep\relax}>{\raggedright\arraybackslash}p{0.3376\dimexpr\textwidth-8\tabcolsep\relax}>{\raggedright\arraybackslash}p{0.2991\dimexpr\textwidth-8\tabcolsep\relax}@{}}
\toprule
\textbf{Block} & \textbf{Field (values)} & \textbf{What it captures} & \textbf{Grounding} \\
\midrule\endfirsthead
\toprule
\textbf{Block} & \textbf{Field (values)} & \textbf{What it captures} & \textbf{Grounding} \\
\midrule\endhead
\bottomrule\endfoot
Task \& domain & Task category (10) & The deliverable that would satisfy the request: drug information; treatment and management; foundational knowledge; patient education; test and result interpretation; documentation and workflow; diagnosis and differential; coding and administrative; procedural guidance; other & Generic clinical-question taxonomies\textsuperscript{44,49,51}; clinical-information-system needs\textsuperscript{50,53}; MedHELM task families\textsuperscript{16}; LLM usage studies\textsuperscript{22,26} \\
 & Question intent (12; 5 groups) & The motive behind the task (verification, fact look-up, clinical decision, dosing, documentation drafting, definition, mechanism, comparison, procedure, coding, result interpretation, other) & Web-search intent\textsuperscript{56}; generic question stems\textsuperscript{44}; well-built clinical question\textsuperscript{57}; question pursuit\textsuperscript{43} \\
 & Clinical department (25); Medicine division (15) & The department (and Medicine division) that would own management, via a precedence ladder & Institutional ownership for governance and oversight (Methods) \\
 & AMA use case (8) & Mapping to the AMA list of physician AI use cases, plus none of the above & AMA physician survey\textsuperscript{1} \\
Query properties & Patient-specific (binary) & About an identifiable patient & Patient-specific needs\textsuperscript{49,53} \\
Context & Patient or local information; current evidence (binary) & Information a safe answer needs that the query lacks & Unmet patient-specific needs\textsuperscript{53}; EHR grounding\textsuperscript{58}; resource gaps\textsuperscript{59}; retrieval augmentation\textsuperscript{29,30} \\
Risk & Potential harm (5: minimal to critical) & Harm if the clinician acted on an incorrect answer & AHRQ Common Formats harm scale\textsuperscript{45,60}; NCC MERP; WHO ICPS; ISMP high-alert medications; pediatric dosing\textsuperscript{61}; older adults\textsuperscript{62} \\
Response & Answerability (3); safest response (5: answer directly, retrieve then answer, clarify, escalate, abstain) & Whether the query is answerable as posed; the least restrictive response that is still safe & Selective prediction and deferral\textsuperscript{32,41,42,63}; retrieval\textsuperscript{29,64}; clarifying questions\textsuperscript{31}; abstention\textsuperscript{65} \\
Surface features & Medication; dose; lab or result; imaging; vulnerable population; acute or urgent; text generation (binary); query form (3); abbreviation (binary) & Literal cues in the text, coded independently of judgments & Medication harm\textsuperscript{61,62}; query logs\textsuperscript{66,67}; abbreviation ambiguity\textsuperscript{68}; documentation tools\textsuperscript{3,4} \\
\end{longtable}
\endgroup
{\small URLs for the AHRQ Common Formats, NCC MERP, WHO ICPS and ISMP sources are given in Methods.\par}
\FloatBarrier

\FloatBarrier\Needspace*{18\baselineskip}\noindent{\small \textbf{Extended Data Table 2 \textbar{} Inter-rater reliability and human--model concordance by field.}\par}\vspace{4pt}
\begingroup\footnotesize\setlength{\tabcolsep}{3pt}\renewcommand{\arraystretch}{1.12}
\begin{longtable}{@{}>{\raggedright\arraybackslash}p{0.2137\dimexpr\textwidth-16\tabcolsep\relax}>{\raggedright\arraybackslash}p{0.1603\dimexpr\textwidth-16\tabcolsep\relax}>{\raggedright\arraybackslash}p{0.0801\dimexpr\textwidth-16\tabcolsep\relax}>{\raggedright\arraybackslash}p{0.1015\dimexpr\textwidth-16\tabcolsep\relax}>{\raggedright\arraybackslash}p{0.0962\dimexpr\textwidth-16\tabcolsep\relax}>{\raggedright\arraybackslash}p{0.1175\dimexpr\textwidth-16\tabcolsep\relax}>{\raggedright\arraybackslash}p{0.1132\dimexpr\textwidth-16\tabcolsep\relax}>{\raggedright\arraybackslash}p{0.1175\dimexpr\textwidth-16\tabcolsep\relax}@{}}
\toprule
\textbf{Field} & \textbf{Krippendorff's \ensuremath{\alpha} (95\% CI)} & \textbf{Gwet's AC1} & \textbf{Pairwise agreement (\%)} & \textbf{Unanimous (\%)} & \textbf{Held-out clinician vs reference (\%)} & \textbf{GPT-5.6 sol vs reference (\%)} & \textbf{Difference, pts (95\% CI)} \\
\midrule\endfirsthead
\toprule
\textbf{Field} & \textbf{Krippendorff's \ensuremath{\alpha} (95\% CI)} & \textbf{Gwet's AC1} & \textbf{Pairwise agreement (\%)} & \textbf{Unanimous (\%)} & \textbf{Held-out clinician vs reference (\%)} & \textbf{GPT-5.6 sol vs reference (\%)} & \textbf{Difference, pts (95\% CI)} \\
\midrule\endhead
\bottomrule\endfoot
\multicolumn{8}{@{}l}{\textbf{Task \& domain}} \\
Task category & 0.64 (0.55--0.73) & 0.67 & 70 & 59 & 84 & 79 & -5 (-14 to 4) \\
Clinical domain & 0.48 (0.38--0.57) & 0.60 & 61 & 48 & 79 & 73 & -6 (-17 to 4) \\
Medicine division & 0.43 (0.33--0.52) & 0.59 & 61 & 45 & 74 & 73 & -2 (-12 to 8) \\
Question intent & 0.45 (0.36--0.54) & 0.51 & 55 & 38 & 69 & 67 & -2 (-16 to 10) \\
Question intent (five groups) & 0.61 (0.50--0.70) & 0.68 & 74 & 62 & 84 & 82 & -2 (-11 to 7) \\
AMA use case & 0.25 (0.14--0.34) & 0.42 & 48 & 26 & 54 & 67 & 13 (1 to 24) \\
\multicolumn{8}{@{}l}{\textbf{Query properties}} \\
Patient-specific & 0.61 (0.45--0.75) & 0.76 & 85 & 78 & 91 & 92 & 0 (-6 to 7) \\
Actionable & 0.39 (0.25--0.51) & 0.44 & 71 & 56 & 79 & 83 & 4 (-5 to 13) \\
Evidence-dependent & 0.29 (0.15--0.42) & 0.33 & 65 & 48 & 73 & 66 & -8 (-19 to 3) \\
\multicolumn{8}{@{}l}{\textbf{Context \& response}} \\
Needs patient context & 0.30 (0.16--0.43) & 0.39 & 67 & 51 & 76 & 84 & 8 (0 to 17) \\
Needs institutional context & 0.25 (0.01--0.49) & 0.89 & 91 & 86 & 95 & 97 & 2 (-2 to 6) \\
Needs current evidence & 0.32 (0.16--0.46) & 0.53 & 72 & 58 & 81 & 86 & 5 (-4 to 14) \\
Needs patient or local information & 0.30 (0.16--0.44) & 0.39 & 67 & 51 & 76 & 84 & 8 (0 to 17) \\
Answerability & 0.57 (0.44--0.68) & 0.49 & 65 & 50 & 77 & 79 & 2 (-8 to 12) \\
Answerability below high & 0.52 (0.38--0.64) & 0.52 & 76 & 64 & 84 & 85 & 1 (-7 to 9) \\
Safest response & 0.40 (0.26--0.52) & 0.49 & 57 & 41 & 72 & 76 & 5 (-5 to 15) \\
\multicolumn{8}{@{}l}{\textbf{Risk}} \\
Risk (5-level) & 0.36 (0.23--0.48) & 0.19 & 34 & 12 & 35 & 48 & 13 (-6 to 34) \\
High or critical risk (binary) & 0.13 (-0.05 to 0.31) & 0.73 & 79 & 69 & 87 & 85 & -2 (-10 to 7) \\
\multicolumn{8}{@{}l}{\textbf{Surface features}} \\
Names a medication & 0.90 (0.82--0.97) & 0.91 & 95 & 93 & 98 & 98 & 0 (-2 to 2) \\
States a dose & 0.85 (0.61--1.00) & 0.98 & 98 & 97 & 99 & 99 & 0 (-2 to 2) \\
Cites a lab/result & 0.64 (0.41--0.81) & 0.91 & 93 & 89 & 96 & 91 & -5 (-11 to 1) \\
Mentions imaging & 0.55 (0.10--0.82) & 0.96 & 96 & 94 & 98 & 98 & -0 (-4 to 3) \\
Vulnerable population & 0.65 (0.15--0.89) & 0.96 & 97 & 95 & 98 & 100 & 1 (0 to 3) \\
Acute or urgent & 0.32 (0.03--0.57) & 0.92 & 93 & 89 & 96 & 98 & 2 (-1 to 4) \\
Requests text generation & 0.74 (0.54--0.88) & 0.93 & 95 & 92 & 97 & 100 & 3 (1 to 5) \\
Query form & 0.71 (0.60--0.80) & 0.75 & 83 & 75 & 91 & 97 & 6 (2 to 11) \\
Uses abbreviation & 0.90 (0.82--0.96) & 0.91 & 95 & 93 & 98 & 99 & 1 (-2 to 4) \\
\end{longtable}
\endgroup
{\small Krippendorff's \ensuremath{\alpha} uses the ordinal metric for harm, answerability and safest response and the nominal metric otherwise; 95\% CIs from 2,000 bootstrap resamples of queries. The reference for concordance is the label shared by two clinicians when they agree; the held-out third clinician and the corpus annotator are scored against the same references. Differences are model minus clinician, with paired bootstrap 95\% CIs (1,000 resamples). The intent groups, patient or local information, answerability below high and high or critical risk are the fields as reported (Methods); actionability and evidence dependence were annotated but are not analyzed.\par}
\FloatBarrier

\FloatBarrier\Needspace*{18\baselineskip}\noindent{\small \textbf{Extended Data Table 3 \textbar{} Logistic regression of high or critical potential harm.}\par}\vspace{4pt}
\begingroup\small\setlength{\tabcolsep}{3pt}\renewcommand{\arraystretch}{1.12}
\begin{longtable}{@{}>{\raggedright\arraybackslash}p{0.4658\dimexpr\textwidth-6\tabcolsep\relax}>{\raggedright\arraybackslash}p{0.2671\dimexpr\textwidth-6\tabcolsep\relax}>{\raggedright\arraybackslash}p{0.2671\dimexpr\textwidth-6\tabcolsep\relax}@{}}
\toprule
\textbf{Variable} & \textbf{Model 1: OR (95\% CI)} & \textbf{Model 2: OR (95\% CI)} \\
\midrule\endfirsthead
\toprule
\textbf{Variable} & \textbf{Model 1: OR (95\% CI)} & \textbf{Model 2: OR (95\% CI)} \\
\midrule\endhead
\bottomrule\endfoot
Concerns a specific patient & 2.91 (2.64--3.22) & 3.20 (2.88--3.55) \\
Names a medication & 3.12 (2.89--3.37) & 1.30 (1.18--1.44) \\
States a dose & 1.13 (0.97--1.31) & 1.93 (1.70--2.18) \\
Cites a laboratory or test result & 0.93 (0.85--1.01) & 0.99 (0.90--1.09) \\
Mentions imaging & 0.62 (0.53--0.73) & 0.76 (0.68--0.86) \\
Vulnerable population & 1.24 (1.08--1.42) & 1.37 (1.24--1.52) \\
Acute or urgent situation & 2.43 (1.98--2.98) & 3.07 (2.52--3.74) \\
Requests text generation & 0.16 (0.11--0.24) & 0.58 (0.44--0.75) \\
Uses clinical abbreviations & 1.43 (1.35--1.51) & 1.87 (1.76--1.98) \\
Query form: command (vs question) & 0.56 (0.41--0.78) & 0.87 (0.68--1.10) \\
Query form: fragment (vs question) & 0.51 (0.46--0.56) & 0.85 (0.79--0.91) \\
Fellow (vs attending) & 1.15 (0.87--1.53) & 1.25 (0.96--1.64) \\
Resident (vs attending) & 1.47 (1.25--1.73) & 1.21 (1.05--1.40) \\
APP (vs attending) & 0.86 (0.73--1.02) & 0.89 (0.76--1.04) \\
Registered nurse (vs attending) & 0.46 (0.40--0.53) & 0.58 (0.50--0.68) \\
Foundational knowledge (vs documentation \& workflow) & --- & 0.18 (0.13--0.24) \\
Drug information \& pharmacotherapy (vs documentation \& workflow) & --- & 15.92 (12.70--19.94) \\
Treatment \& management (vs documentation \& workflow) & --- & 19.48 (15.41--24.62) \\
Diagnosis \& differential (vs documentation \& workflow) & --- & 7.77 (6.25--9.66) \\
Test \& result interpretation (vs documentation \& workflow) & --- & 3.37 (2.67--4.25) \\
Patient education \& communication (vs documentation \& workflow) & --- & 1.68 (1.24--2.28) \\
Coding \& administrative (vs documentation \& workflow) & --- & 0.09 (0.04--0.21) \\
Procedural guidance (vs documentation \& workflow) & --- & 29.23 (22.16--38.55) \\
Other (vs documentation \& workflow) & --- & 0.41 (0.33--0.50) \\
\end{longtable}
\endgroup
{\small Odds ratios with 95\% CIs from standard errors clustered by clinician; n = 127,625 queries from 6,342 clinicians. Model 1 includes the surface features, patient specificity, query form and professional family; model 2 adds task category (reference, documentation and workflow).\par}
\FloatBarrier

\end{document}